%% file: paper-main.tex
\documentclass[acmsmall]{acmart}
\usepackage{xspace}
\usepackage{xcolor}
\usepackage{enumitem}
\usepackage{newunicodechar}
\newunicodechar{∗}{\textasteriskcentered}

\AtBeginDocument{%
  }

\setcopyright{cc}
\setcctype{by}
\acmJournal{PACMHCI}
\acmYear{2025} \acmVolume{9} \acmNumber{7} \acmArticle{CSCW330} \acmMonth{11} \acmPrice{}\acmDOI{10.1145/3757511}

\begin{document}

\title{Time is On My Side: Dynamics of Talk-Time Sharing in Video-chat Conversations}

\author{Kaixiang Zhang}
\email{kz88@cornell.edu}
\affiliation{%
  \institution{Cornell University}
  \city{Ithaca}
  \state{New York}
  \country{USA}
}

\author{Justine Zhang}
\email{tisjune@umich.edu}
\affiliation{%
  \institution{University of Michigan}
  \city{Ann Arbor}
  \state{Michigan}
  \country{USA}
}

\author{Cristian Danescu-Niculescu-Mizil}
\email{cristian@cs.cornell.edu}
\affiliation{%
  \institution{Cornell University}
  \city{Ithaca}
  \state{New York}
  \country{USA}
}

\renewcommand{\shortauthors}{Kaixiang Zhang, Justine Zhang, and Cristian Danescu-Niculescu-Mizil}

\input{macros}

\begin{abstract}
\input{000abstract}
\end{abstract}

\ccsdesc[500]{Human-centered computing~Collaborative and social computing design and evaluation methods}

\keywords{conversations; interaction patterns}

\maketitle

\section{Introduction}
\label{sec:intro}

\input{010intro}

\section{Background}
\label{sec:related}

\input{020related}

\section{Main Conversational Setting}
\label{sec:data}
\input{030data-v2}

\section{A Framework for Analyzing the Sharing of Talk-Time}
\label{sec:method}
\input{040method}
\input{050typology}

\minorrevision{
\section{Case Studies}
\label{sec:cases}
\input{110casestudies}
\subsection{Exploratory analysis: Speaker characteristics and time sharing dynamics} \label{sec:additional}
}

\input{060additional}
\minorrevision{
\subsection{Application to Supreme Court Oral Arguments}\label{sec:extension}
}
\input{100extension}

\section{Discussion}
\label{sec:discussion}
\input{070conclusions}

\begin{acks}
\input{120acknowledgements}
\end{acks}

\bibliographystyle{ACM-Reference-Format}
\bibliography{refs}

\appendix

\section{Appendix}
\label{sec:appendix}
\input{090appendixsection}

\received{July 2024}
\received[revised]{December 2024}
\received[accepted]{March 2025}

\end{document}

%% file: macros.tex
\newif\ifshowcomments
\showcommentsfalse
\newcommand\todo{\textcolor{orange}}
\newcommand{\jz}[1]{{\textcolor{teal}{JZ:#1}}}
\newcommand{\cd}[1]{\textcolor{blue}{CD:#1}}
\newcommand{\kz}[1]{{\textcolor{magenta}{KZ:#1}}}

\newcommand{\revision}[1]{#1} %
\newcommand{\internal}[1]{#1} %
\newcommand{\minorrevision}[1]{#1} %
\newcommand{\wordingrevision}[1]{#1}

\newcommand{\cut}[1]{}
\newcommand{\xhdr}[1]{{\noindent\bfseries #1.}}

\newcommand\TT{talk-time\xspace}
\newcommand\STTlong{sharing of \TT\xspace}
\newcommand\TTsharing{\TT sharing\xspace}

\newcommand\overallbalance{conversation-level balance\xspace}
\newcommand\overallimbalance{conversation-level imbalance\xspace}

\newcommand\balance{balance\xspace}
\newcommand\balanced{balanced\xspace}
\newcommand\highlybalanced{highly-\balanced\xspace}
\newcommand\imbalance{imbalance\xspace}
\newcommand\imbalanced{imbalanced\xspace}
\newcommand\highlyimbalanced{highly-\imbalanced\xspace}
\newcommand\primary{primary\xspace}
\newcommand\secondary{secondary\xspace}
\newcommand\windowimbalance{window \imbalance}

\newcommand\conversationwindow{conversation window}
\newcommand\interleave{interleave\xspace}
\newcommand\blue{\textcolor{blue}{blue}\xspace}
\newcommand\red{\textcolor{red}{red}\xspace}
\newcommand\gray{\textcolor{gray}{gray}\xspace}

%% file: 000abstract.tex
An intrinsic aspect of every conversation is the way talk-time is shared between multiple speakers.
Conversations can be balanced, with each speaker claiming a similar amount of talk-time, or imbalanced when one 
talks disproportionately.
Such overall distributions are the consequence of 
continuous negotiations between the speakers throughout the conversation: 
who should be talking at every point in time, and for how long?

In this work we introduce a computational framework for quantifying both the conversation-level distribution of talk-time between speakers, as well as the lower-level dynamics that lead to it.
We derive a typology of talk-time sharing dynamics structured by several intuitive axes of variation.
By applying this framework to a large dataset of video-chats between strangers, 
we confirm that, perhaps unsurprisingly, different conversation-level distributions of talk-time are perceived differently by 
speakers, with balanced conversations being preferred over imbalanced ones, especially by those who end up 
talking
less.
Then we reveal that---even when they lead to the same level of overall balance---different types of talk-time sharing dynamics are perceived differently by the participants, highlighting the relevance of our newly introduced typology.
Finally, we discuss how our framework offers new tools to designers of computer-mediated communication platforms, for both human-human and human-AI communication. 

%% file: 010intro.tex
In a conversation, talk must be somehow distributed among the speakers.
Some people talk more than others, perhaps remarkably so---thus the complaint that one is being talked over, or the worry that one has talked too much.
In a classic study of talk at a dinner party, \citet{tannen_conversational_2005} examines many aspects of the guests' conduct; among them, the "high-involvement" behaviors of some guests---who talk a lot and leave little room for interjection---lead others to report feeling "dominated."
Tannen's analysis additionally traces how such feelings of fitting in or getting left out are 
\wordingrevision{produced},
as the dinner conversation moves from back-and-forth banter to focused attention on a single storyteller.  
In this way, the distribution---and constant renegotiation---of talk-time in a conversation can meaningfully shape the speakers' experiences.

In this paper, we explore how talk-time is distributed between speakers
in video-chat conversations.
We go beyond thinking of this distribution as a static, conversation-level characteristic---i.e., the extent to which some speakers talk more than others---and instead conceive of it as the result of a \textit{dynamic} process that develops throughout the entire duration of the interaction.
As we will show, attending to the dynamics is important for revealing potentially consequential distinctions.
A perfectly balanced conversation can be comprised of rapid-fire back-and-forth exchanges or of alternating monologues; accordingly, a broad sense of interactional equity may be modulated by differing experiences of flow or engagement.

We propose a computational framework to examine the dynamics of \TTsharing.
Our approach builds off of a simple measurement: for a given span of conversation, we quantify the fraction of time taken by each speaker. 
Going span-by-span, we represent 
each 
conversation in terms of how the distribution of talk-time changes 
as the conversation progresses.
Based on these representations, we then 
derive a space of possible talk-time sharing dynamics structured by several intuitive axes of variation.

To demonstrate its use, we apply our framework to a collection of over 1,500 video-chat conversations between pairs of strangers \cite{reece_candor_2023}.
We examine how different 
types of
\TTsharing dynamics correspond to speakers' post-hoc assessments of how their conversations went.
Perhaps unsurprisingly,
we find that conversations that are overall more balanced in terms of the distribution of talk-time
are preferred.
However, we reveal striking differences in speakers’ perceptions when comparing different dynamics that lead to such balanced conversations.
In particular, balanced conversations 
\wordingrevision{where}
speakers engage in back-and-forth interactions are perceived as being significantly less enjoyable than those 
\wordingrevision{where}
speakers alternate in dominating the talk-time.
Such distinctions, alongside others we observe, highlight the need for not only analyzing static snapshots of the conversation-level distribution of talk-time, but also the lower-level dynamics that lead to them.

We additionally use our framework 
\wordingrevision{to perform}
an exploratory analysis of the relation between talk-time sharing dynamics and speakers' characteristics.
We find that speakers are consistent in their talk-time sharing behavior across multiple conversations, and that this consistency can in part be explained by age and gender demographics.
Although exploratory and limited to a single domain, our analyses 
\wordingrevision{suggest potential ways for}
systematic inequities in the way speakers from different groups interact with each other 
\wordingrevision{to show up in conversations.}

\wordingrevision{We also show how our framework can be used to surface meaningful conversation dynamics in other domains. 
While our analysis primarily focuses on the video-chat data, as an illustrative example, we apply the framework to a collection of Supreme Court Oral Arguments---a scenario that presents notable contrasts to the video-chat setting, since it involves multiple speakers with pre-determined roles interacting face-to-face.
To encourage broader applications, we distribute the code that implements our framework, together with demonstrations on the video-chat and Supreme Court datasets.}

From a practical standpoint, we argue that researchers should attend to the dynamics of talk-time sharing---beyond static qualities like the degree of conversation-level balance---when analyzing or designing for computer-mediated conversations.
\revision{
We offer our general framework as a 
tool for examining such dynamics 
and extending a design space of possible interventions that could potentially encourage more equitable 
\citep{leshed_visualizing_2009,do_how_2022,kim_meeting_2008,viegas_chat_1999,lam_wpclubhouse_2011}
and enjoyable~\citep{guydish_pursuit_2023} interactions.}
In \wordingrevision{light} of \wordingrevision{the} increasing prominence of human-AI interactions, our framework also surfaces an additional dimension to consider when assessing the (in)capability of AI agents to have human-like conversations~\citep{zhang_ideal_2021,reeves_conversational_2017,stokoe_elizabeth_2024,dingemanse_text_2022}.

From a theoretical standpoint,
our paper considers
a central preoccupation of existing scholarship on interaction: how conversation happens \citep[inter alia]{sidnell_handbook_2012,schegloff_lectures_1992,yeomans_practical_2023,dingemanse_text_2022}.
A wide range of past work in fields like linguistics and sociology has focused on such mechanics as taking turns \citep{sacks_simplest_1974,levinson_turn-taking_2016,stivers_universals_2009}, 
telling stories \citep{tannen_conversational_2005,jefferson_sequential_1978,mandelbaum_storytelling_2012}, and bringing conversations to a close \citep{schegloff_opening_1973}.
Our framework 
\wordingrevision{contributes} a novel perspective on the dynamic distribution of talk,
and a way of analyzing large-scale conversational corpora through that lens.
In particular, while past studies have examined the interplay between regimes of talk where one speaker leads versus where both equally contribute \citep{dingemanse_text_2022,gilmartin_chats_2018,eggins_analysing_2004},
we offer a way of holistically modeling conversations in terms of such regimes so that they can be systematically compared to each other.

%% file: 020related.tex
Our work focuses on the mechanics of how conversation happens.
In this way, we share similar interests as past work from fields like psychology, sociology, and linguistics that examines 
how people take turns~\citep{sacks_simplest_1974,levinson_turn-taking_2016,stivers_universals_2009},
sequence actions~\citep{schegloff_sequence_2007,clark_polite_1980},
address communicative troubles~\citep{schegloff_preference_1977},
engage in interactional modes like storytelling~\citep{jefferson_sequential_1978,cook-gumperz_social_2006,bavelas_listeners_2000,dingemanse_text_2022},
or start or end conversations \citep{schegloff_routine_1986,schegloff_opening_1973}---alongside numerous other aspects.
Some work more explicitly engages with what people say---e.g., 
what feelings are expressed
or what outcomes are arrived at (see \citet{yeomans_practical_2023} for a survey).
Such factors are somewhat out of scope for our study, 
though we draw on them when interpreting our dynamics of interest.
Rather, we engage with the presumption---shared by a diverse range of paradigms from conversation analysis \citep{hoey_conversation_2017,schegloff_lectures_1992} to computational linguistics \citep{grosz_attention_1986,grosz_centering_1994,dingemanse_text_2022}---that it's worth accounting for the things people do that make conversation (dis)orderly or (in)coherent.
Conversation analysis in particular attends to how coherence is \textit{achieved} by speakers as an interaction progresses \citep{schegloff_lectures_1992,hoey_conversation_2017};
guided by this sensibility, we make the \textit{constant renegotiation} of talk-time distribution a central focus of our approach.

One recurring concern in 
such research
is how talk is allocated.
Numerous studies focus on turn-taking:
who speaks next, and when do they start \citep{levinson_turn-taking_2016,sacks_simplest_1974,schegloff_overlapping_2000}? 
In investigating how speakers ``orient to'' the norm of ``one speaker at a time,'' some work also considers violations of this norm, 
like interruptions \citep{drummond_backward_1989,goldberg_interrupting_1990}
or silences \citep{kurzon_towards_2007,roberts_interaction_2006,heldner_pauses_2010}.
Other work goes beyond individual turns to examine 
who has the floor~\citep{edelsky_whos_1981,goffman_forms_1981},
who's in control~\citep{walker_mixed_1990,jefferson_caveat_1993,nguyen_modeling_2014},
or who's at the center of attention~\citep{grosz_centering_1994,grosz_attention_1986}.
This perspective abstracts away from ambiguities relating to what counts as a turn:
an individual utterance could be construed as an interruption, a backchannel, or a meaningful change in who's speaking \citep{levinson_turn-taking_2016,goffman_forms_1981}.

Our work adds to these accounts of the distribution of talk.
We consider broader spans of conversation, as opposed to individual utterances or turns,
placing our conceptualization closer to those about control or floor.
Our work is not the first to do this---other accounts have drawn a distinction between ``chunks'' (where one speaker takes the floor and dominates for an extended period) and ``chats'' (where the floor may be jointly developed), documenting their distributions and the ways they transition between one another \citep{edelsky_whos_1981, dingemanse_text_2022,eggins_analysing_2004,gilmartin_chats_2018}.
However, we go on to propose ways of comparing between different types of conversations, based on these characteristics. 
This allows us to go beyond corpus-level claims about the composition of talk---where such studies have generally focused---to provide a systematic account of the diversity of conversational dynamics within a corpus.

Our computational approach and our level of abstraction present certain compromises.
Our framework should not be interpreted as an accurate model of ideas like control or floor---which concern speakers' subjective impressions~\citep{edelsky_whos_1981,goffman_forms_1981}---so much as a means \wordingrevision{of} methodically \wordingrevision{surfacing} interactional patterns that could point to these ideas.
Our coarse focus on talk-time 
ignores the intricate and contextual interpretations of detail that are
characteristic of approaches like conversation analysis \citep{schegloff_reflections_1993,hoey_conversation_2017,sidnell_handbook_2012}.
Instead, 
following other writing on the complementary perspectives offered by quantitative and qualitative methods (e.g., \citep{baker_useful_2008,pardo-guerra_extended_2022}),
we see our approach as providing a birds-eye view over a space of conversational dynamics 
(see \citep{zhang_characterizing_2018} for an analogue in a social-media setting).
By systematically applying a simple measure---share of talk-time---across 
subsequent spans of interaction, we are able to produce holistic representations of conversations, 
organize them in an intuitively structured space,
and make claims about their 
relative
prevalence in particular 
contexts.
As we discuss in Section \ref{sec:discussion}, there are many ways for future, finer-grained approaches to elaborate on the scaffolding we've set up.

\xhdr{Relation to speakers and their experiences}
How do different ways of allocating talk relate to speakers' impressions of a conversation?
Past studies often approach this question in terms of whether or not interactions are balanced, or whether certain people contribute larger shares of talk.
Across multiple settings, researchers have drawn connections between balance and 
\wordingrevision{other factors like task performance}
\citep{pardo_montclair_2019,niculae_conversational_2016,cao_my_2021} 
and group dynamics \citep{mast_dominance_2002,fay_group_2000,maclaren_testing_2020,prabhakaran_who_2013}.
Numerous studies have also examined the relationship between gender and the distribution of talk \citep{brescoll_who_2011,mast_dominance_2002,james_understanding_1993,bilous_dominance_1988}, 
often drawing connections to the unequal levels of power conventionally held by men versus women (e.g., \citep{brescoll_who_2011}).

Past work generally conceives of balance as a good thing, especially when speakers are on equal footing: 
per \citet{guydish_reciprocity_2021} and \citet{guydish_pursuit_2023}, conversational balance could reflect \wordingrevision{that} interlocutors 
\wordingrevision{have} complementary, reciprocal behaviors, 
and could result in decreased social distance.
Per \citet{lakoff_logic_1973} and \citet{tannen_conversational_2005}, dominating the conversation could be seen as imposing and impolite,
\wordingrevision{while,}
as Tannen's dinner party account shows, less-active speakers could feel excluded in an imbalanced interaction.

Some other work has attended to speakers' perceptions of interactional patterns beyond the \overallbalance of participation.
For instance, some studies have tied fast-paced back-and-forths, 
or a preponderance of interruptions to social connection or ``clicking''~\citep{templeton_fast_2022,mcfarland_making_2013}.
Others have examined regimes of talk indicative of storytelling \citep{bavelas_listeners_2000,jefferson_talking_2015,mandelbaum_storytelling_2012,jefferson_sequential_1978,eggins_analysing_2004} or self-disclosure practices \citep{sprecher_taking_2013},
suggesting emotionally and socially beneficial effects (e.g., \citep{sprecher_taking_2013}; see also \citep{labov_telling_2010}).
Our approach provides a way of investigating similar questions and extending such analyses to more 
fine-grained
models of talk-time sharing.
We illustrate this in our empirical case study, detailed in Sections~\ref{sec:method}~and~\ref{sec:additional}.

\xhdr{Talk-time considerations in computer-mediated settings}
A broad range of research on computer-mediated interaction has focused on conversational dynamics in such settings. 
Our work is most directly relevant to studies of \textit{synchronous} communication, including face-to-face interactions, 
audio calls, 
and video chat:
the real-time nature of such settings renders the question of how participants share talk-time especially pertinent.
However, as we discuss in Section \ref{sec:discussion}, the distribution of talk remains pertinent for other channels such as 
instant messaging.

Several past studies have proposed ways of improving interactions (e.g., \citep{seering_designing_2019}).
Some of these approaches have focused on mediating synchronous conversations,
proposing interfaces or automated tools aimed at making speakers aware of their conversational practices (e.g., \citep{leshed_visualizing_2009}).
Of particular relevance to us, many of these interventions are concerned, at least in part, with whether the relative degree to which different people participate is balanced or skewed~\citep{leshed_visualizing_2009,do_how_2022,kim_meeting_2008}.
Other studies have documented gender-based imbalances in contributions to collaborative online efforts like Wikipedia~\citep{lam_wpclubhouse_2011,gallus_gender_2020},
in the service of making participation in such projects more equitable.
These studies have mostly focused on analyzing and modulating overall levels of imbalance in interactions;
our work surfaces additional dimensions---relating to the dynamics of talk---that designers and maintainers of such systems could consider.

A large body of work has also focused on designing conversational agents that can interact with people in realistic ways;
\wordingrevision{such a} project 
\wordingrevision{has} gained prominence with recent developments in generative AI.
Some scholars have drawn on accounts of conversational practices to examine chatbot behaviors,
suggesting ways that these agents could be improved, or specifying new criteria for their evaluation \citep{reeves_conversational_2017,stokoe_elizabeth_2024,dingemanse_text_2022}.
\wordingrevision{Our framework, in providing an additional perspective on conversational dynamics, adds to these efforts.}

%% file: 030data-v2.tex
To concretize the discussion of our approach, we start by describing the \wordingrevision{primary} dataset we apply our framework to.
We examine CANDOR, a large corpus of video chat dialogs \citep{reece_candor_2023}, 
in which speakers were paired with people they'd never met, and simply told to have a conversation with no further stipulations.
In using the framework to analyze this data, we will illustrate the types of conversational dynamics it can capture;
to make sense of the characterizations our method produces, we will draw on survey responses from the speakers in these conversations.

The distribution of talk-time is relevant in any setting where multiple people interact, 
but for the purpose of an \textit{initial} exploration, the CANDOR collection has several advantages. 
First, there are no pre-determined speaker roles. 
In more asymmetric scenarios---consider teacher-student dialogs---there would be additional
constraints or norms governing how much talk each person takes up \citep{heritage_talk_1992,clayman_watershed_2010}.
No such a-priori expectations are applicable here, 
so we can more fully explore the possible space of talk-time sharing dynamics.
Second, the dialogs are synchronous.
In asynchronous settings,
the distribution of talk-time would be impacted not just by the contingencies of real-time interaction,
but also by absences as people close chat windows or email clients (e.g., \citep{kuutti_when_2003}).
This would introduce extra types of complexities when interpreting what the conversational dynamics mean.
Finally, the conversations in the CANDOR corpus are dyadic.
As such, our framework needs only to make a single distinction between the relative contributions of the two speakers, 
rather than deal with more complex configurations \citep{simmel_number_1902,stivers_is_2021},
such as subgroups having side chats.

\revision{
While analyzing talk-time sharing dynamics in dyadic video-chat conversations is our main focus in this work, we also test how our framework can be used in a different scenario.  
In Section~\ref{sec:extension}, we discuss its application to a face-to-face, multi-speaker setting with asymmetric relations: the Supreme Court Oral Arguments.  
We further discuss factors that limit the generalizability of our framework and suggest avenues for addressing them in Section~\ref{sec:discussion}.}

\xhdr{Data description}
The CANDOR corpus consists of 1,656 conversations involving 1,456 unique speakers, 
and was collected by other researchers to facilitate large-scale studies of naturalistic conversation.
The speakers, all located in the United States, were randomly paired with each other and instructed to have a 25-minute conversation, after which they'd be compensated.
Both the audio and video of the conversation were recorded, and transcriptions were automatically produced using the AWS Transcribe API.
\begin{figure}[h]
  \centering
  \includegraphics[width=0.55\textwidth]{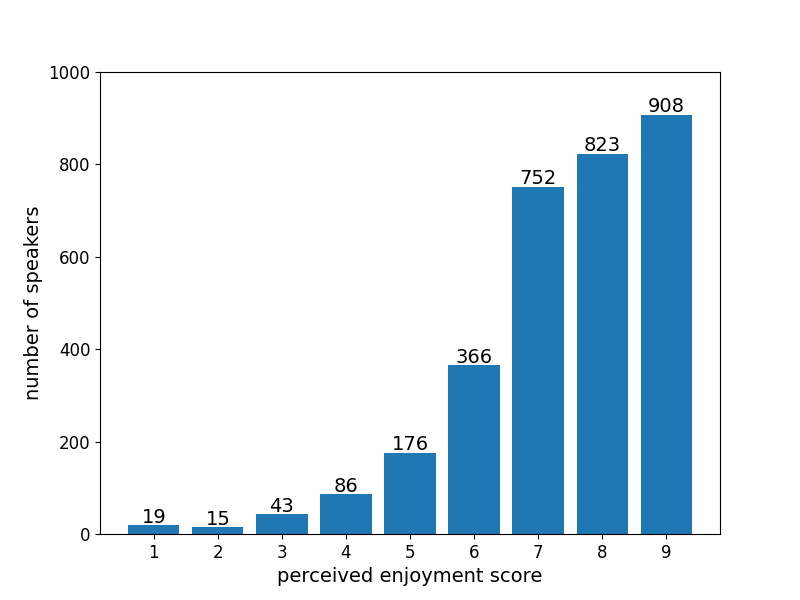}
  \caption{Distribution of enjoyment rating among all speakers from the CANDOR corpus.  Most speakers rate their conversations very highly, with 28.5\% giving the maximum rating.} 
  \Description{Distribution of enjoyment rating among all speakers for all conversations from the CANDOR corpus. Bar plot showing a strongly right-skewed distribution.}
  \label{fig:candor_enjoy_distribution_all}
\end{figure}

Speakers also completed surveys about their conversational experiences. 
We focus on two survey responses in particular.
Speakers provided a numerical score of \textbf{how enjoyable} they perceived the conversation to be. 
As Figure~\ref{fig:candor_enjoy_distribution_all} shows, these scores have a heavy positive skew. 
As such, when measuring the reported enjoyment over a set of conversations, we use the percentage of conversations that obtained the maximum score of 9,
in addition to taking 
the mean enjoyment score.
Speakers also wrote comments in which they separately discussed the \textbf{positive and negative aspects} of the interaction;
examples are provided in Table~\ref{tab:comment_example}.
\begin{table*}
  \caption{Example post-conversation comments for a conversation in the CANDOR corpus (id: 36f338b1-bb0e-4c79-b40d-1c584913f262). 
  Speaker A reported an enjoyment score of 9---the maximum value, while Speaker B reported an enjoyment score of 5.
  }
  \label{tab:comment_example}
  {\small
    \begin{tabular}{c|p{10cm}}
        \toprule
         & Example post-conversation comments \\ \hline
         Speaker A & 
        \textbf{Positive:} Everything went well, my partner was very pleasant, sociable, positive, warm, I actually learnt some new things during our conversation and I felt like I wanted to extend our conversation for a longer time and I think the feeling was mutual. I do believe my partner enjoyed the conversation too. \newline
        \textbf{Negative:} There's nothing that want wrong during the conversation with my partner. My partner and I had a great time and we really enjoyed our conversation, we were both happy and actually it looked to me that if I didn't mention about the fact that our time was up my partner would have continued the conversation which means the conversation was really enjoyable and natural.\\ \hline
         Speaker B & 
        \textbf{Positive:} [NAME] was very conversational and had a lot of ideas for things to chat about. We had some things in common that allowed us to have deeper conversation than surface level. She was good at asking questions  and allowing space for the answers, and appeared interested in what I was saying. \newline
        \textbf{Negative:} [NAME] made space for me to talk and asked a lot of questions, but she spoke very quickly and didn't make much time for me to ask her questions. She also didn't seem to listen to my answers very specifically. She didn't allow much of a two way street in the conversation.\\
        \bottomrule
    \end{tabular}
    } %
\end{table*}

We use these responses in complementary ways.
If our framework makes a distinction between two sets of conversations with differing conversational dynamics,
then differences in enjoyment score between these two sets would indicate that the dynamics we model have some bearing on speakers' experience---they contribute to different impressions of what happened, or co-occur with other interactional phenomena that produce different levels of enjoyment.
The written comments can then provide us with additional clues about the types of experiences these dynamics point to.
While we find these surveys extremely rich, we note that other paradigms---notably conversation analysis---would draw on what people said in the conversation, 
rather than their post-hoc assessments \citep{hoey_conversation_2017,sidnell_handbook_2012} (see also \citep{stafford_conversational_1987}); 
we discuss how our computational approach could be extended to do this in Section~\ref{sec:discussion}.

For the sake of examining conversations that are comparable to each other, we exclude conversations where speakers reported significant technical issues like dropped calls,
or where speakers didn't complete the 
\wordingrevision{study,}
i.e., who had particularly short conversations, or who didn't complete the survey. 
This results in 
1,594 conversations for our analysis.

\xhdr{Measuring talk-time}
Our framework starts from quantifying the amount of time each speaker spends talking.
Transcripts in the CANDOR corpus consist of utterances from speakers and their start and end timestamps; 
we take the duration of an utterance to be its end time minus its start time,
and the talk-time of a speaker to be the duration of all their utterances.
\wordingrevision{Since the transcripts are generated by an automated transcription system, which may introduce errors, we validate the speaking times derived from the transcripts against those obtained by directly processing the audio files.}\footnote{We also re-run all the analyses starting directly from the audio signal, and obtain qualitatively similar results. Details of these comparisons are included in Appendix~\ref{sec:appendix-audio}.}
We note that audio provides an alternative, perhaps richer, way to determine talk-time (e.g., \citep{heldner_pauses_2010}), which future work could explore.

%% file: 040method.tex
Our framework for examining the distribution of talk-time operates at two scales.
We start with a straightforward, conversation-level measurement: to what extent is there an \textit{imbalance} in the time taken by speakers?
At a 
lower-level, we then consider how this imbalance dynamically changes as the conversation progresses.
We structure the ensuing discussion accordingly: first, we introduce a 
simple
measure of imbalance and explore how it relates to speakers' impressions of the conversation;
then, we introduce a procedure to track and systematically analyze the dynamics of talk-time sharing, and identify different types of interactional patterns that---while reflective of equal levels of \overallimbalance---lead to diverging perceptions of the conversation.
We make our framework’s implementation publicly available as part of ConvoKit \cite{chang_convokit_2020}, enabling others to extend it and adapt it to other conversational settings. \footnote{https://convokit.cornell.edu/}

\subsection{Preliminary measure: conversation-level imbalance}
\label{sec:imbalance}
Our framework builds off a simple characterization of talk-time sharing: a conversation is balanced when all speakers talk for a similar amount of time and is imbalanced when some speakers take up more time than others.
In what follows, we'll describe our framework for two-person conversations.
Formally, we measure \textit{imbalance} as the fraction of total talk-time in a conversation that's taken up by the most talkative speaker. 
We refer to the more talkative speaker as the \textit{primary speaker} and the less talkative speaker as the \textit{secondary speaker}.

\begin{figure}[h]
  \centering
  \includegraphics[width=0.56\linewidth]{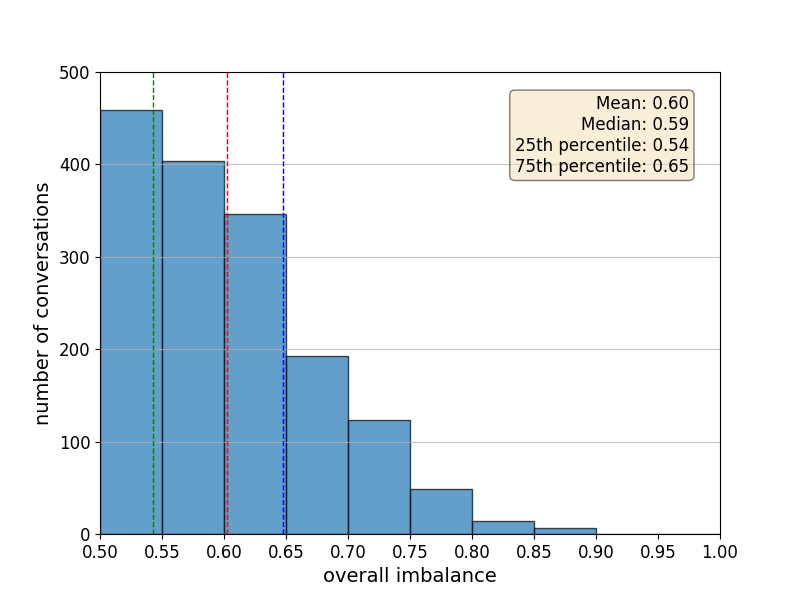}
  \caption{Distribution of \overallimbalance in the CANDOR corpus.}
  \Description{Plot showing the distribution of the degree of imbalance for conversations across the CANDOR corpus. Bar plot with decreasing number of conversations as degree of imbalance increase. The mean, median, 25th percentile, and 75th percentile, in that order, are 0.63, 0.62, 0.55, and 0.69. The max is reaching 0.95 and the low is 0.5.}
  \label{fig:candor_balance_distribution}
\end{figure}
Figure~\ref{fig:candor_balance_distribution} shows the distribution of imbalance values over the CANDOR corpus.
We note that many conversations are fairly balanced between the two speakers, (imbalance~$\approx \frac{1}{2}$).
However, for a substantial portion of the data,
one speaker talks over twice as much as the other (imbalance~$\geq \frac{2}{3}$).

\xhdr{Speakers' perception}
While basic, we see that our imbalance measure points to diverging conversational experiences;
perhaps unsurprisingly (e.g., \citep{guydish_pursuit_2023,guydish_reciprocity_2021}), people generally prefer balanced interactions over imbalanced ones.
Comparing between \textit{highly-balanced} conversations (top quartile of imbalance score) and \textit{highly-imbalanced} conversations (bottom quartile)
we see that in highly-balanced conversations \internal{31.8\%} of the speakers give a maximum enjoyment score, versus only \internal{25.3\%} in highly-imbalanced conversations.
The respective mean enjoyment scores given by speakers in the two types of conversations are 7.53 and \internal{7.25}, respectively; 
these differences are statistically significant according to a Mann-Whitney U test ($p < 0.001$).

To further make sense of this distinction, we examine speakers' comments: 
what sorts of remarks are more likely to be associated with balanced versus imbalanced conversations?
To compare comments across different parts of the data, we use the ``Fightin' Words'' statistical approach from \citet{monroe_fightin_2008} that quantifies the extent to which a phrase (up to three words) occurs more frequently in comments from one part of the data (e.g., highly-balanced conversations) versus the other (e.g., highly-imbalanced conversations), relative to a word-frequency distribution bias.\footnote{Comparing to the prior---here, an uninformative Dirichlet prior---accounts for statistical irregularities stemming from overall word frequencies.}
This results in a list of phrases that are particularly distinctive in one versus the other part of the data (in our case, in highly-balanced versus highly-imbalanced conversations).

\begin{table*}
  \caption{Phrases that most distinguish between \balanced and \imbalanced conversations, and representative excerpts including them. The comparisons are done separately for comments discussing positive and negative aspects. \revision{Additional examples are included in Table~\ref{tab:candor_fw_balance_imbalance_extended_a} and ~\ref{tab:candor_fw_balance_imbalance_extended_b} in the Appendix.}}
  \label{tab:candor_fw_balance_imbalance}
  {\small
  \begin{tabular}{p{2.6cm}|p{3cm}|p{7.2cm}}
    \toprule
            & top distinctive phrases & example excerpts \\ \hline
    Positive~comments~for \textbf{\highlybalanced} \newline conversations & 
    both, lot in common, were able, were able to, like we, lot in, about our, able, able to, we both &
    \noindent \textbullet \textbf{We both} had a \textbf{lot in common}, so it made talking with each other super easy, and almost \textbf{like we} knew each other. \newline
    \textbullet\ I feel \textbf{like we} connected and that makes it easier. \newline
    \textbullet\ I felt open and \textbf{able to} share what is going on in my city and life. \\ \hline
    
    Positive comments for \textbf{\highlyimbalanced} conversations & 
    chat, he, partner, talked lot, listener, else, im, is, she, bring &
    \noindent \textbullet\ I feel like my \textbf{chat partner} was great at asking me questions and keeping me talking. \newline
    \textbullet\ \textbf{She talked a lot} and was very pleasent and smiled a lot. \newline
    \textbullet\ Fortunately, I am a good \textbf{listener}. \\
    \midrule\midrule
    
    Negative~comments~for \textbf{\highlybalanced} \newline  conversations & 
    our, our conversation, think of, of anything that, we had, anything that, both, her and, think the conversation, each other &
    \noindent \textbullet\ We flowed easily in \textbf{our conversation} and I am typing to get to fifty word count. \newline
    \textbullet\ I think the rough parts were mainly how we tried to \textbf{think of} how to go back and forth, at first. \newline
    \textbullet\ We \textbf{both} tend to be more introverted as we mentioned in the conversation. \\ \hline
    
    Negative~comments for \textbf{\highlyimbalanced} conversations & 
    than, more than, much, talked, me, too much, talked too, talked too much, he, ended up, ended &
    \noindent \textbullet\ \textbf{He talked} a lot \textbf{more than} I did. \newline
    \textbullet\ I probably \textbf{talked too much}, but I felt like she wasn't giving very much in return so I don't think I had a choice. \newline
    \textbullet\ \textbf{He ended up} opening up toward the end and actually provided me with some information that I might find useful for everyday life as well. \\
  \bottomrule
\end{tabular}
} %
\end{table*}
\revision{Table~\ref{tab:candor_fw_balance_imbalance} contains distinguishing phrases for highly-balanced and highly-imbalanced conversations, along with comment excerpts that contain them (additional examples in Table~\ref{tab:candor_fw_balance_imbalance_extended_a} and ~\ref{tab:candor_fw_balance_imbalance_extended_b} in the Appendix).} 
For a more interpretable analysis, we use the Fightin' Words method to compare positive comments and negative comments separately.
We see that speakers from highly-balanced conversations are more likely to 
refer
to the joint aspects of the conversation: the most distinguishing words include ``both'', ``we both'', ``our'', ``each other'', and ``lot in common'', perhaps suggesting connections between balance and other forms of discursive convergence \citep{guydish_pursuit_2023}. 
On the other hand, speakers from \highlyimbalanced conversations are more likely to refer to individuals (``he'', ``she'', ``me'', ``i'm'', and  ``partner''),
or to the skewed distribution of talk (``talked too much'').
We find this joint-vs-individualistic distinction 
in both the positive and negative comments.

\begin{figure}[h]
  \centering
  \includegraphics[width=0.65\linewidth]{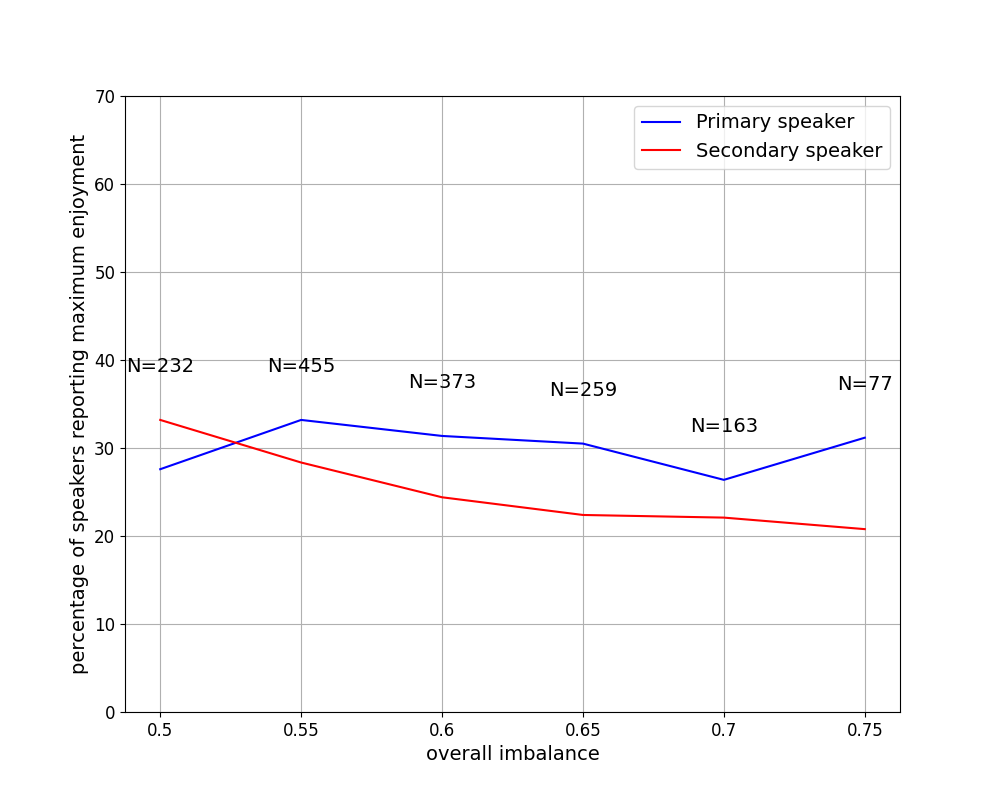}
  \caption{Percentage of \primary and \secondary speakers reporting maximum enjoyment scores for conversations with different levels of \imbalance.  The number of conversations for each level is indicated as $N$. Levels with 30 or less conversations ($N < 30$) are not shown.}
  \Description{We show percentage of high enjoyment score received from primary and secondary speakers of conversations at different degrees of \imbalance. One blue line representing primary speaker's enjoyment and one red line representing secondary speaker's enjoyment at varying degree of \imbalance for conversations. The two lines start near each other, both going downward and develop larger gap as conversation get more \imbalance.}
  \label{fig:candor_enjoy_vs_balance}
\end{figure}
\xhdr{Difference between \primary and \secondary speakers}
These patterns confirm that speakers often take notice imbalance of talk-time, and take actions or form judgements that reflect this characteristic of the conversation. 
However, do both speakers react to \imbalance similarly?  
This question is especially pertinent in \highlyimbalanced conversations where the roles of \primary and \secondary speaker become more salient.
Figure~\ref{fig:candor_enjoy_vs_balance} shows the relation between perceived enjoyment and level of imbalance for each type of speaker.
We notice that the more the conversation is \imbalanced, the less likely it is for the \secondary speaker to enjoy it.
This leads to an increasing gap between the way the \primary and \secondary speakers perceive their conversation.

\revision{
This gap is also reflected when comparing the post-conversation comments of \primary and \secondary speaker in highly-imbalanced conversations, again using the Fightin' Words method (Table~\ref{tab:candor_fw_primary_secondary}, extended in Table~\ref{tab:candor_fw_primary_secondary_extended_a} and ~\ref{tab:candor_fw_primary_secondary_extended_b} inåthe Appendix).

Here, we find intuitive distinctions in the primary and secondary speakers' characterizations of how much their conversation partners talked.
Additionally, we reveal a dichotomy on how 
\wordingrevision{talkativeness}
(or lack thereof) is perceived.
Secondary speakers can perceive the talkativeness of their partners positively, appreciating that they ``had [a] lot of'' interesting ``stories'' to ``share''.
But they can also perceive it as detrimental to the conversation flow: ``it was hard'' to contribute when the other ``talked [a] lot''.
This dichotomy is mirrored in the perceptions of the primary speakers: they can either feel connected when their partner is a ``good listener'' who shows ``interest'' by ``asking questions'',  or they can regret ``talk[ing] too much'' and perceive the other person as being too ``shy''.
}%

\revision{
\wordingrevision{Such divergences in the perception of talk-time imbalance
can result from different mechanisms, 
as prior work on floor-taking and conversational control suggests \cite{edelsky_whos_1981,palmer_controlling_1989,brescoll_who_2011,nguyen_modeling_2014}.
A speaker can talk more because they have a lot to say or because their interlocutor isn't contributing;
similarly a speaker can talk less because they don't have opportunities to do so or because they don't take those opportunities when they arise.
Complementing our talk-time framework with analyses of utterance content could help disentangle the mechanisms at play,
and in turn motivate design strategies for improving conversational dynamics.} 

}%

\begin{table*}
  \caption{Phrases that distinguish most between comments by \primary speakers and \secondary speakers for highly-\imbalanced conversations, and representative excerpts including them. The comparisons are done separately for positive comments and for negative comments. \revision{Additional examples are included in Table~\ref{tab:candor_fw_primary_secondary_extended_a} and ~\ref{tab:candor_fw_primary_secondary_extended_b} in the Appendix.}}
  \label{tab:candor_fw_primary_secondary}
  {\small
  \begin{tabular}{p{2.6cm}|p{3cm}|p{7.2cm}}
    \toprule
            & top distinctive phrases & example excerpts \\ \hline
    Positive comments \newline from \textbf{\primary speaker} & 
    questions, asked, but, probably, us, about my, agreed, some, young, good listener, each, interested &
    \noindent \textbullet She was a \textbf{good listener} and \textbf{asked} pertinent \textbf{questions} to get me talking. \newline %
    \textbullet\ I probably did the majority of the talking, but they \textbf{asked} good \textbf{questions} and gave interesting answers when I \textbf{asked} them \textbf{questions}. \newline %
    \textbullet\ She was also fun to talk to because she's a \textbf{good listener} and was understanding toward my situation. \\ \hline %

    Positive comments \newline from~\textbf{\secondary speaker} & 
    had lot, something, lot, stories, had similar, had lot of, share, friendly and, that my, life &
    \noindent \textbullet He \textbf{had a lot} to talk about so I let him tell me about those things for as long as he wanted to. \newline %
    \textbullet\ He had great life experience and \textbf{stories} that he \textbf{shared}. \newline %
    \textbullet\ My partner was very chatty and willing to \textbf{share} her views, while I was quite willing to listen. \\
    \midrule\midrule

    Negative comments \newline from \textbf{\primary speaker} & 
    her, little, too much, we had, flow, talked too, should, she was, she is, shy &
    \noindent %
    \textbullet\ I think \textbf{she was} a \textbf{little} quiet by nature which is ok. \newline %
    \textbullet\ I think I probably \textbf{talked too much}. \newline %
    \textbullet\ I felt I led the conversation \textbf{too much} and perhaps dominated. \\ \hline %

    Negative comments \newline from \textbf{\secondary speaker} & 
    different, hard, say, went, the conversation was, talked lot, life, honestly, didnt, kept, it was hard &
    \noindent %
    \textbullet\ He talked way too much; \textbf{it was hard} to get a word in edgewise. \newline %
    \textbullet\ I did not know what to \textbf{say}. \newline %
    \textbullet\ My partner \textbf{talked a lot} about herself and \textbf{didn't} ask many questions, which made it harder for me to share as well. \\ %
  \bottomrule
\end{tabular}
} %
\end{table*}

%% file: 050typology.tex
\subsection{Main framework: \TTsharing dynamics}
\label{sec:typology}

Conversation-level imbalance is a consequence of continuous negotiations between speakers at every point of the interaction.
At any moment, a speaker might switch from taking the lead to focusing on the other speaker, or engage in a back-and-forth, sharing time more equally with their partner.
Here, we introduce a method to capture and systematically examine such dynamics. 
Following our approach, we end up with a \textit{structured space} of talk-time sharing dynamics that 
can offer new insights into conversational data.

One basic element of our formalism is a \textit{\conversationwindow}: an arbitrarily small span of the conversation covering a specific length of time.
We take a sliding window approach to capture the way the conversation progresses: each conversation is composed of multiple windows of $K$ seconds, \internal{sliding in increments of $L$ seconds (and thus overlapping for $K-L$ seconds)}, where $L<K$.\footnote{For our analysis, we use $K=150s$ and $L=30s$, arriving at these parameters after initial exploration of the CANDOR data; minor variations of $K$ and $L$ lead to similar results.} 
\wordingrevision{For each such \conversationwindow, we measure the \imbalance of \TT, i.e., the fraction of talk-time taken up by the most talkative speaker \textit{within that window} (as opposed to across the entire conversation, as in Section~\ref{sec:imbalance}).}
This simple approach allows us to capture the dynamics of \TTsharing: 
\wordingrevision{at what moments is \TT dominated by one speaker versus the other,
at what moments is \TT equally distributed,
and in what sequence do these different regimes occur?}

\begin{figure}[h]
  \centering
  \includegraphics[width=0.7\linewidth]{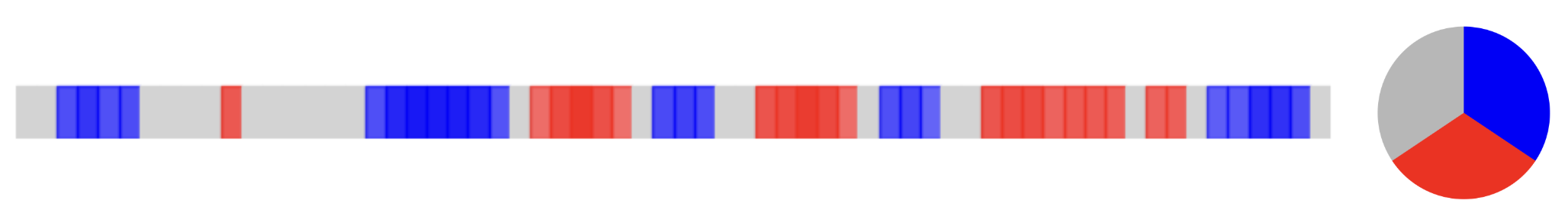}
  \caption{Visualization of window-by-window talk-time sharing dynamics for an individual conversation
 (CANDOR ID: a7b14ca1-0b36-42b9-ad4a-50f0eb094035). The conversation level imbalance for this conversation is \internal{0.51}.}
  \Description{Shows one visualization plot for a individual conversation based on the method proposed from the method section. 
  The plot is a single horizontal rectangle with blue, red, and gray coloring on each small block representing windows following the visualization method. A pie chart is also presented, with blue, red, and gray colors reflecting percentage of blue, red, and gray color from the visualization.}
  \label{fig:candor_individual_plot}
\end{figure}

To aid our explanation, we introduce a way to visualize the sequence of talk-time sharing regimes in a given conversation (following the example in Figure \ref{fig:candor_individual_plot}).
We use three colors to indicate the three respective regimes for a window: \blue for windows where the (eventual conversation-level) \primary speaker talks the most, \red for windows where the (conversation-level) \secondary speaker talks the most, and \gray for windows in which the two speakers talk a similar amount of time (i.e., neither speaker's talk time 
\wordingrevision{exceeds}
$M\%$ of the total window \TT; in our analysis on the CANDOR corpus we set $M=60\%$).\footnote{We note that such distinctions are similar to those made in \citet{edelsky_whos_1981} between singly- and jointly-developed floors, and in \citet{eggins_analysing_2004} and \citet{gilmartin_chats_2018} between ``chunk'' and ``chat'' sequences.}
The shades of \blue and \red correspond to the percentage of the window \TT the respective speaker talks, with darker shades indicating higher percentages.
The color-coded windows are illustrated in succession, without overlap between subsequent windows for clarity (even though their actual time overlaps).
Finally, a pie-chart offers an overview by showing the percentage of \blue, \red, and \gray windows; note that by construction, the percentage of \blue windows will most often always be larger than that of \red windows, because by construction \blue corresponds to the speaker \wordingrevision{who}
talk the most overall.

This representation, as illustrated in Figure~\ref{fig:candor_individual_plot}, reveals some aspects of the conversation that go beyond the conversation-level view offered by the simple measure of conversation-level imbalance described earlier.
First, the overall primary speaker doesn't necessarily dominate the entire conversation, as shown in the presence of \gray and \red windows alongside the \blue ones.
Second, we see how the conversation progresses through multiple regimes of talk-time sharing: after some initial greetings and introductions (first \gray window), the primary speaker takes the lead for most of the first half of the conversation \internal{(mostly \blue and \gray windows)}, narrating their mental health struggles during the initial months of the COVID-19 pandemic; the conversation then moves to a stretch led by the secondary speaker, who takes up the discussion of mental health to talk about their childrens' difficulties with school (long stretches of \red windows).

\xhdr{A structured space of talk-time sharing dynamics}
Together, the regimes depicted in Figure~\ref{fig:candor_individual_plot} amount to a fairly balanced conversation (overall conversation-level imbalance \internal{$=0.51$}).
However, in attending to the local dynamics of talk, we note that a different composition of regimes could have also resulted in the same \overallimbalance.
For instance, a different, similarly-balanced conversation could consist of speakers who engage in more back-and-forths (i.e., more \gray windows), rather than switching between longer stretches where one person leads.

We now propose a way of systematically making such distinctions, by mapping these 
representations to a structured space.
We start by identifying three extreme cases:
the primary speaker can \textbf{dominate throughout} a conversation, visualized as mostly \blue windows;
both speakers can solely engage in balanced \textbf{back-and-forth} exchanges, visualized as mostly \gray windows;
or, the secondary speaker can dominate in as many windows as possible---in such conversations, the primary and secondary speaker \textbf{alternate in dominating} the conversation, visualized as a mixture of \blue and \red windows 
(recall that by construction, the primary speaker almost always dominates more windows than the secondary, placing an upper limit on the possible proportion of \red windows).

\begin{figure}[h]
  \centering
  \includegraphics[width=0.9\linewidth]{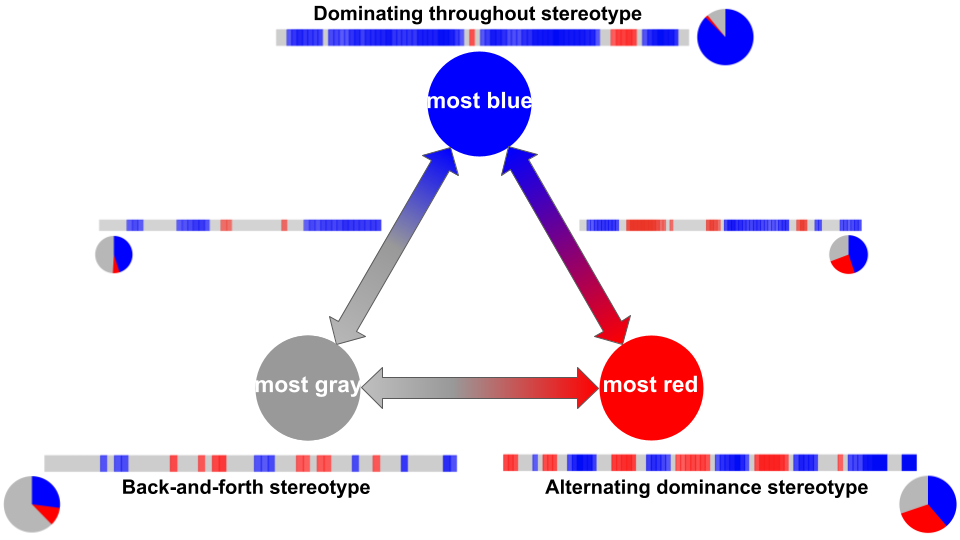}
  \caption{A structured space of \TT dynamics.
  The extremes are represented by \blue, \red and \gray circles, and are accompanied by visualizations of example stereotypical conversations. 
  \revision{Arrows indicate axes along which conversations can vary between each stereotype, and two example conversations that fall along these axes in between the stereotypes are included.}
  Figure~\ref{fig:candor_enjoy_triangle} shows the difference in reported enjoyment for conversations that lie at the extremes of the axes and 
  Figure~\ref{fig:candor_fix_blue_enjoy} shows how reported enjoyment varies along the bottom axis; 
  variation along the other two axes are illustrated in the Appendix.}
  \Description{Visualizing our proposed triangle typology for categorizing conversations from micro-scale. The top of the triangle is conversation with most blue, representing the dominating throughout type that are conversations dominated by blue windows (by primary speakers). Down to the bottom of the triangle conversation gets separated into back and forth and \interleave type conversations, based on if the rest of the conversation windows other than blue are filled with gray or red. The two vertices on the bottom is one gray and one red, showing the extreme conversations with most possible gray windows and red windows in terms of percentage.}
  \label{fig:candor_triangle}
\end{figure}

As depicted in Figure~\ref{fig:candor_triangle}, these three conversational ``stereotypes'' bound a space of possible conversational dynamics.
Of course, most conversations do not resemble the extreme cases,
but by moving between these points, we can systematically account for richer interactional patterns.
For instance,
alternating periods of talk-time dominance can be interrupted by more equally-distributed back-and-forths;
such dynamics correspond to conversations that lie on the axis between the ``most \gray'' and ``most \red'' extremes.

\xhdr{Speakers' perceptions}
\wordingrevision{Starting from the three conversational stereotypes, we organize and systematically investigate the space of \TTsharing dynamics.}
In particular, we note that while the top of the space in Figure \ref{fig:candor_triangle} (most \blue) corresponds to very imbalanced conversations, the bottom (most \gray, most \red, and the space in-between) correspond to different ways of achieving balance in \wordingrevision{a} conversation.
Therefore, we can ask whether the way \TT balance is achieved---either through back-and-forths or through alternating dominance---lead to different speaker perceptions, and thus refining our previous analysis that grouped all balanced conversations together.

\begin{figure}[h]
  \centering
  \includegraphics[width=0.6\linewidth]{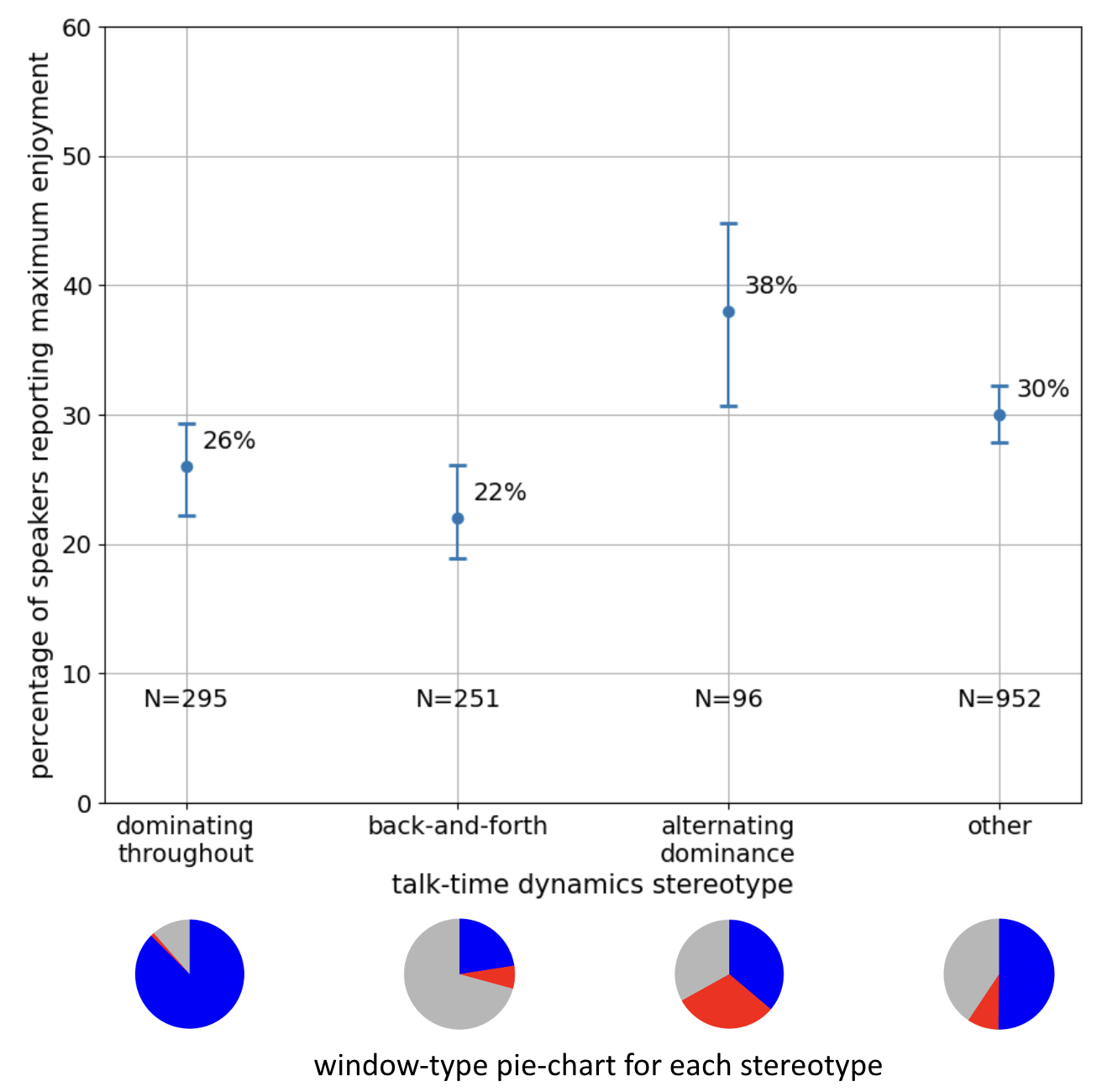}
  \caption{Comparing the reported enjoyment of the three stereotypical \TT dynamics: dominating throughout, back-and-forth, and alternating dominance; for reference, we include conversations that do not fit in any of these stereotypes (other).
  We show the percentage of maximum enjoyment scores and 95\% confidence intervals obtained via bootstrap resampling. 
  Pie charts are included to show the average percentages of window regime types for each stereotype.
  }
  \Description{Percentage of speakers reporting maximum enjoyment is highest for alternating dominance type conversations, followed by dominating throughout and back-and-forth is lowest. The error bar was larger for alternating dominance and back-and-forth due to less datapoints, but the order remains the same. Pie charts are also presented fro each type, with mostly blue for dominating throughout, mostly gray for back-and-forth, and almost equal for all three colors for alternating dominance.}
  \label{fig:candor_enjoy_triangle}
\end{figure}
Figure~\ref{fig:candor_enjoy_triangle} shows the difference in perceived enjoyment for conversations that lie at the extremes of the axes in the CANDOR corpus.
We see that even though stereotypical back-and-forth conversations (more than 60\% \gray windows)\footnote{We define boundary percentages heuristically, based on data distributions.} and stereotypical alternating dominance conversations (more than 25\% \red windows) have relatively similar conversation-level \imbalance (0.54 and 0.53 average conversation-level imbalance, respectively), there is a strong preference for the alternating dominance dynamics (average enjoyment score of \internal{{7.79} vs. {7.26}} for back-and-forth, $p<0.0001$ according to a Mann Whitney U test).\footnote{To check that the observed difference \wordingrevision{in enjoyment} is not explained by the difference in conversation-level imbalance,
we also make a controlled comparison between back-and-forth and alternating dominance stereotypical conversations that are paired by their conversation-level imbalance (with tolerance of \internal{0.005}). 
We obtain \internal{93} pairs which show no difference in conversation-level balance (Wilcoxon signed-rank test p-value = \internal{$0.42$}).  In this tightly controlled set, we still observe a substantial difference in perceived enjoyment (\internal{$p< 0.05$}).}

In fact, back-and-forth conversations have very similar levels of enjoyment as the much more imbalanced dominated throughout conversations, \wordingrevision{which we take to be those with} more than 75\% \blue windows (average \overallimbalance of \internal{0.72}; average enjoyment score of {7.2}; Mann Whitney U test gives \internal{$p = 0.86$} comparing mean enjoyment between back-and-forth and dominating throughout conversations).

\revision{
To further interpret the distinction between back-and-forth and alternating dominance conversations, we examine distinguishing phrases in the post-conversation comments, shown in Table~\ref{tab:candor_fw_alternate_backforth} and Table~\ref{tab:candor_fw_alternate_backforth_extended_a} and ~\ref{tab:candor_fw_alternate_backforth_extended_b} in the Appendix.
Comparing positive comments, we note a stark difference in experience: while speakers in back-and-forth conversations are \wordingrevision{often} only able to ``keep the conversation going'' by finding some common ``topics'', those engaging in alternating dominance were \wordingrevision{more likely to be} interested in ``listen[ing]'' to ``the other's'' ``stories'' and ``life experiences''.
These two distinct ways of connecting with a stranger also came with their specific challenges. 
\wordingrevision{Many of the} negative comments for back-and-forth conversations 
\wordingrevision{pointed to a necessary effortfulness: participants}
``had to'' do something to ``keep'' the conversation ``flow'' and avoid awkward ``moments''.  
The negative comments for alternating dominance 
\wordingrevision{were more diverse, though many questioned}
how genuine the interlocutor's interest actually was.
In summary, the differences in the talk-time sharing dynamics of balanced conversations 
\wordingrevision{pointed to distinctions}
between familiar modes of interaction---in this setting, sharing extended stories or self-disclosures (e.g., \citep{jefferson_sequential_1978,mandelbaum_storytelling_2012,sprecher_taking_2013}) versus (perhaps awkwardly) searching for conversation topics \cite{jurafsky_extracting_2009}.
}%

\begin{table*}
  \caption{Most distinguishing phrases between comments for stereotypical alternating dominance and back-and-forth conversations, and representative excerpts including them. The comparisons are done separately for positive comments and for negative comments. \revision{Additional examples are included in Table~\ref{tab:candor_fw_alternate_backforth_extended_a} and ~\ref{tab:candor_fw_alternate_backforth_extended_b} in the Appendix.}}
  \label{tab:candor_fw_alternate_backforth}
  {\small
  \begin{tabular}{p{3.2cm}|p{3cm}|p{6.5cm}}
    \toprule
            & top distinctive phrases & example excerpts \\ \hline
    Positive~comments~for \textbf{alternating dominance} conversations & 
    stories, to listen, listen, stranger, life experiences, the other, myself, his, good conversation, super &
    \noindent \textbullet\ We both had interesting \textbf{stories} to tell, so that was also very enjoyable. \newline %
    \textbullet\ I feel like we were both able \textbf{to listen} actively to one another, and bring curiosity to the conversation. \newline %
    \textbullet\ We both listened when \textbf{the other} was speaking and asked follow-up questions. \\ \hline %
    
    Positive comments for \textbf{back-and-forth} conversations & 
    which, the conversation going, conversation going, topics, keep the, keep the conversation, games, going, well was, discussed &
    \noindent \textbullet\ We both took turns talking and \textbf{keeping the conversation going} with each other. \newline %
    \textbullet\ The flow of the conversation went well in which we kept bringing up \textbf{topics} to speak on and it went smoothly.  \newline %
    \textbullet\ Not a lot outside of talking about \textbf{games}. \\
    \midrule\midrule

    Negative comments for \textbf{alternating dominance} conversations & 
    go, one, interested, she, into the, minutes, get, wasnt really, through, few minutes &
    \noindent %
    \textbullet\ My partner didn't seem \textbf{interested} in me at all. \newline %
    \textbullet\ As soon as we were a \textbf{few minutes} over, \textbf{she} wanted to end the conversation. \newline %
    \textbullet\ \textbf{She} laughed at stuff that \textbf{wasn't really} funny several times, so I couldn't tell if it was for humor or nerves. \\ \hline %
    Negative comments for \textbf{back-and-forth} conversations & 
    some, going, im, most, had to, went well, understand, flow, kept, topics, moments &
    \noindent     \textbullet\ The conversation had \textbf{some} awkward \textbf{moments} and silent pauses. \newline %
    \textbullet\ He didn't have a lot of talking points and I felt as though I \textbf{had to} keep asking questions to keep the conversation \textbf{going}. \newline %
    \textbullet\ The conversation \textbf{flow} seemed forced especially when one topic had been discussed and we need to find another. \\
    \bottomrule
\end{tabular}
} %
\end{table*}

\begin{figure}[h]
  \centering
  \includegraphics[width=0.6\linewidth]{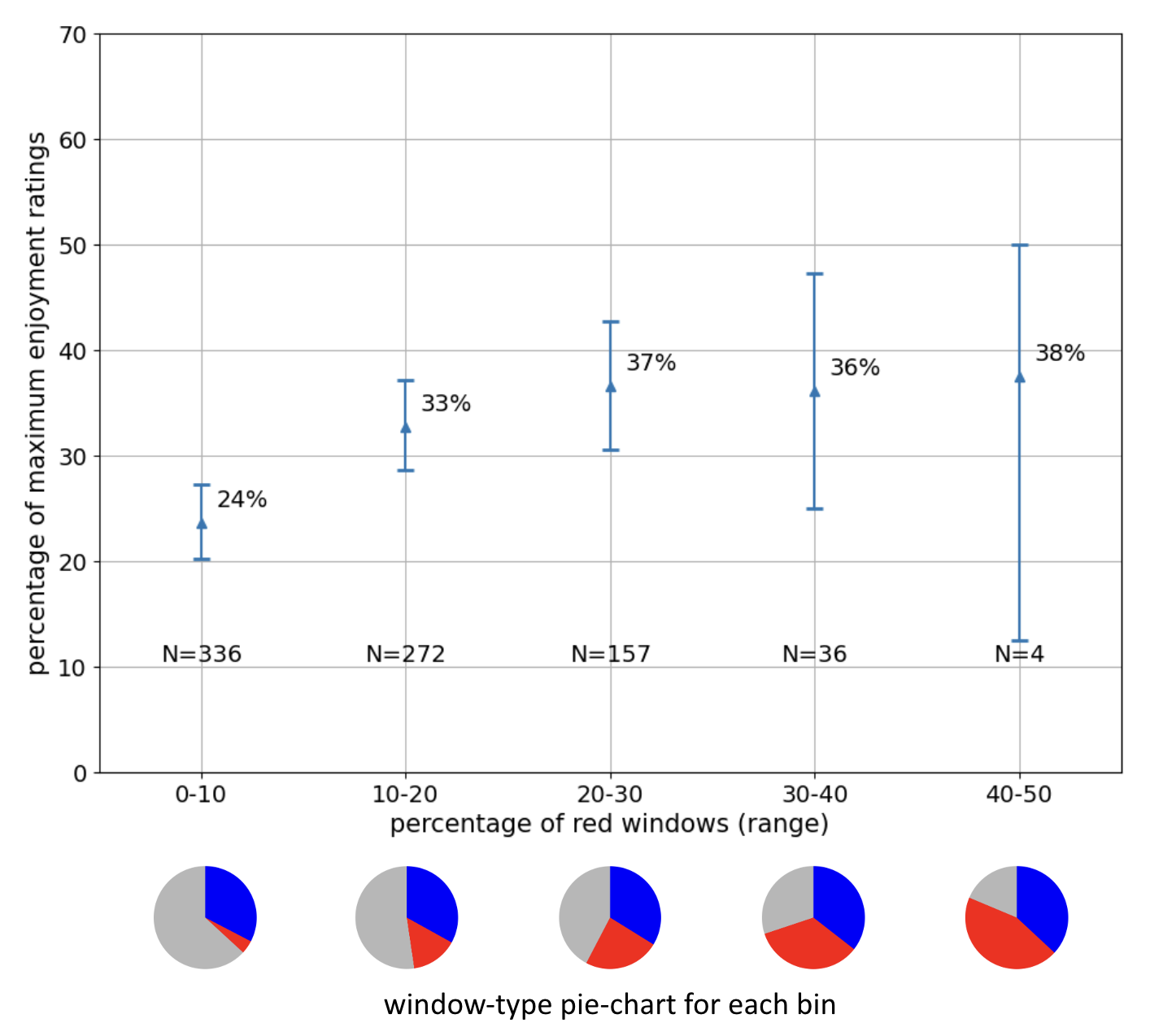}
  \caption{After controlling for the proportion of \blue windows, we compare reported enjoyment for varying proportions of \red windows. 
  }
  \Description{We show in plot, with almost fixed percentage of windows dominated by primary speaker (blue) across all levels, the more windows dominated by secondary speaker (red), the higher enjoyment perceived by participants.}
  \label{fig:candor_fix_blue_enjoy}
\end{figure}

To extend our study beyond stereotypical conversations, we consider the variation across the \gray-\red axis.
Here, we examine conversations with varying proportions of \red windows.
\wordingrevision{To separate this analysis from our preceding discussion of conversation-level imbalance, we analyze conversations where the percentage of \blue windows is below 50\% (the ``bottom'' of the space illustrated in Figure \ref{fig:candor_triangle}).}
Figure~\ref{fig:candor_fix_blue_enjoy} shows that
\wordingrevision{reported enjoyment increases}
as the proportion of \red windows increases---i.e., 
\wordingrevision{the secondary speaker leads}
for longer periods of time.\footnote{Variation in enjoyment scores along the other two axes are depicted in 
the Appendix: variation across the \blue - \red axis yields differences in speaker perception that match with the decrease in conversation-level imbalance, variation across the \blue - \gray axis however does not show any significant differences in speaker perception, in spite of also corresponding to a decrease in conversation-level imbalance. Comparing these two ways of going from imbalanced conversations to balanced conversations further justifies accounting for finer-grained dynamics.}

\xhdr{Changes in dominant speaker}
We explore one further (even finer-grained) type of variation that our framework surfaces: we differentiate between conversations with alternating dominance dynamics based on the number of ``alternations,'' i.e., the number of times the dominant speaker role \textit{flips} between the primary and secondary speaker.
At one extreme, a conversation could contain only one such flip, i.e., a sequence of \blue windows followed by a sequence of \red windows.
At the other extreme, there could be many such flips; for instance, the conversation depicted in Figure~\ref{fig:candor_individual_plot} has \internal{8} flips.\footnote{Different stretches of \blue or \red windows could be punctuated by \gray windows throughout.
For this analysis we ignore all \gray windows and compute \blue-\red or \red-\blue transitions after removing \gray windows.}
Even if they have the same overall levels of participation, these two scenarios seem different, at least at an intuitive level:
single-flip conversations seem more akin to two monologues stitched together,
while speakers trade off the leading role in multi-flip scenarios, 
making the conversation potentially more dialogic or reciprocal.
In order to examine flips while controlling for the relative proportion of different regimes (i.e., as depicted in the pie charts),
we match each single-flip conversation with one that has 3 or more flips, such that they have the same proportion of \blue, \red, and \gray windows (with a tolerance of \internal{2\%}).
This matching process leads to a small number (\internal{$N=90$}) of tightly controlled pairs.
In \internal{53.3\%} of these pairs, the multi-flip conversations receive higher aggregate enjoyment scores,\footnote{To obtain a conversation-level aggregate enjoyment scores, we add the enjoyment scores from each speaker.} 
compared to only \internal{37.8\%} of pairs where the single-flip conversations receive higher scores (in the rest of the cases the scores are equal); \internal{$p=0.06$ according to a one-tailed sign test}.

\begin{figure}[h]
  \centering
  \includegraphics[width=0.7\linewidth]{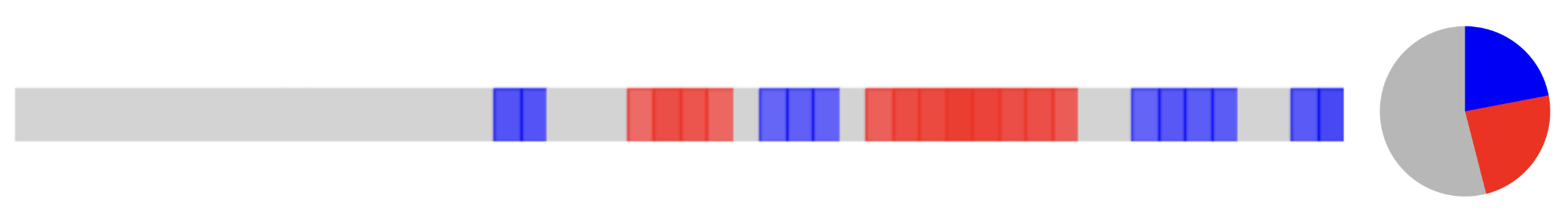}
  \caption{\revision{Visualization of a conversation with talk-time sharing dynamics stereotype transitioning from back-and-forth to alternating dominance (CANDOR ID: 6fb27106-8840-406e-a800-99d1516191ce). 
  Additional examples of conversations with mixed regimes are included in Figures~\ref{fig:bf-dt}, \ref{fig:bf-ad}, \ref{fig:ad-bf}, \ref{fig:ad-dt}, and \ref{fig:dt-ad} in the Appendix.}}
  \Description{Shows one visualization plot for a conversation which the first half is all gray, and the second half is a mix of blue and red.}
  \label{fig:combine_type_convo}
\end{figure}

\xhdr{Mixed dynamics}
\wordingrevision{
Conversations can also transition between different talk-time sharing regimes:
for instance, the conversation visualized in Figure~\ref{fig:combine_type_convo} starts with a back-and-forth dynamic before transitioning to an alternating dominance regime.
Accounting for this heterogeneity is straightforward: 
we apply our framework to different parts of the conversation.
For this analysis, we focus on capturing when conversations switch from one regime to another partway through,\footnote{We note that this analysis can easily be extended to account for mixtures of more than two regimes.}
by using our framework to characterize the first 60\% and the last 60\% of each conversation. \footnote{We use 60\%, allowing overlap between the two halves of the conversation, rather than a strict 50\% cut, to account for the transition in between, enhancing robustness in capturing gradual shifts.}
We find that the most common transitions in this setting involved going from a back-and-forth to alternating dominance regime, and the other way around 
(13 conversations captured for each transition direction, additional examples in Figure \ref{fig:bf-ad} and Figure \ref{fig:ad-bf}),
while mixes involving the dominating-throughout regime are relatively rare (Figures \ref{fig:ad-dt} and \ref{fig:dt-ad}).
}

%% file: 110casestudies.tex
To further illustrate the utility of our framework, we present two case studies.
First, we conduct an exploratory analysis on how speaker characteristics relate to talk-time sharing dynamics in the CANDOR corpus. 
Second, we apply our framework to a dataset of Supreme Court Oral Arguments---comprising structured, multi-speaker interactions---to demonstrate the framework's applicability to a contrasting setting.
Together, these applications underscore the framework’s extensibility and generalizability.

%% file: 060additional.tex
\revision{
Here, we illustrate how our framework can be used to examine the tendencies of speakers with different characteristics to engage in different talk-time sharing dynamics.
\wordingrevision{Rather than trying to make broad claims about the relation between social identity and interaction style based on the CANDOR corpus, we present this analysis as an example of another potential application of the framework that can tie into existing scholarship.} 
}

Past work has suggested connections between speakers' characteristics---including gender and cultural background \citep[inter alia]{zimmerman_9_1996,gumperz_discourse_1982,tannen_indirectness_1981,brescoll_who_2011}---and conversational behaviors.
\citet{tannen_conversational_2005}, for instance, posits that the behaviors of guests at the dinner party she analyzes may reflect the contexts in which they were socialized, and points to \textit{conversational styles} that pervade across different interactions.
In the CANDOR corpus, we see that speakers themselves link their conversational experiences to assessments of their own, or their partners', character traits (e.g., ``introvert,'' ``shy,'' ``talkative'').
Besides contributing to the existing literature on psychological and cultural determinants of conversational dynamics,
an investigation of how talk-time sharing relates to speakers' attributes could also inform designers interested in promoting better, or more equitable \wordingrevision{conversations}
(see Section \ref{sec:discussion} for further discussion).
We approach this in two ways:
we examine whether speakers' roles are consistent across their conversations;
we then analyze how these roles, and the interactional patterns they take part in, are related to demographic factors.

\xhdr{Role consistency across conversations}
Do speakers' characteristics like ``shyness'' or ``talkativeness'' pervade across the conversations they take part in?
We address this question in terms of whether people are consistent in \wordingrevision{being} the primary or secondary speaker, 
making use of the subset of participants in the CANDOR study who take part in multiple conversations.
For the \internal{148} speakers who participate in multiple highly-imbalanced conversations,
we randomly select two conversations and compare their role in them.
We find that \internal{88.1\%} of these speakers
are either the primary speaker in both conversations, or the secondary speaker in both---a much higher rate than agreement by chance (Cohen's \internal{$\kappa=0.76$}).
This suggests that in this setting, there is a strong tendency for speakers to stick to the same role across multiple conversations, even with different conversation partners. 

\xhdr{Relation to demographic attributes}
What speaker characteristics might relate to their conversational roles?
We focus on gender and age, two attributes that speakers self-report in their survey responses.\footnote{In terms of gender, speakers can report either ``Male'', ``Female'', ``Other'', or ``prefer not to answer''.}
Over the \internal{781} conversations involving a male speaker paired with a female speaker,
 we see that in 56\% of them, the primary speaker is male ($p < 0.001$ per a two-tailed signed test),
consistent with several past studies demonstrating the propensity of male speakers to dominate interactions.
Over the 1,295 conversations where speakers differed in age by at least 3 years, 
we see that in 65\% of them, the primary speaker is the older of the pair ($p < 0.001$ per a two-tailed signed test).

\revision{As noted above, these analyses presented here are exploratory and are meant to demonstrate potential uses of our framework, rather than making general distinctions between demographics.
Additional information and data would be needed to determine what accounts for these findings, and whether they are specific to the CANDOR setting, especially given the complex nature of social identity \citep{sen_race_2016}---for instance, 
to what extent do differences in conversational dynamics reflect the speakers, versus their partners' \textit{perceptions} of them \citep{smith_effects_1975,page_effect_1978}?}
Nonetheless, this investigation provides examples of how our framework can be applied to start addressing such questions. 

%% file: 100extension.tex
While the main focus in this work is on 
the video-chat dataset,
\wordingrevision{our framework can---with some adaptation---be used to examine a diversity of other settings.}
As a demonstration,
we apply our framework to a collection of Supreme Court Oral Arguments \cite{shapiro_oral_1984}.
\wordingrevision{This domain differs from our primary video-chat setting in several key ways, allowing us to illustrate the generalizability of our framework.
In contrast to a casual, dyadic exchange involving two strangers on equal footing, there are more than two speakers occupying different institutional roles: lawyers argue for or against a case, while justices listen to these arguments and ask questions.
This has clear implications for speakers' expected share of talk-time;
lawyers almost always speak much more than justices.}
Finally, oral arguments are generally face-to-face as opposed to 
mediated by a video-chat platform.

\xhdr{Data description}
The Supreme Court Oral Arguments dataset \cite{danescu-niculescu-mizil_echoes_2012,chang_convokit_2020} 
\wordingrevision{contains transcripts of oral arguments for cases from the US Supreme Court dating back to 1955.}
Each argument involves nine Supreme Court justices and lawyers from two sides---representing the petitioner and the respondent, respectively---who take turns to present their arguments and answer the justices' questions.
For brevity, here we only consider the petitioner's turn and discuss the results for the respondent's turn in the Appendix (Section \ref{sec:supreme-respondent}).
We filter out cases with missing or additional roles from any party, resulting in 4,528 Supreme Court cases for our analysis.
The dataset also includes 
professional transcripts with time-stamps, which we use to determine talk-time.

The data also includes the nine justices' votes for each case, which are cast after the oral arguments.
\wordingrevision{For this demonstration, we examine whether the distribution of votes is related to the oral argument discussion dynamics.}
\wordingrevision{Here, we focus on comparing discussions that preceded a unanimous vote---where all justices align---with those that preceded a divided vote---which is split 5-to-4.}
In the data, there are 1,760 unanimous votes and 816 divided ones.

\xhdr{Adapting the framework}
To use our framework in this setting, we first need to account for the fact that these interactions have more than two speakers.
\wordingrevision{Here, we treat all the lawyers from one side as a single speaker, as well as all of the justices;
as such, during the petitioners' turn, there are effectively two speakers: the lawyers and the justices.
Future work could explore more elaborate adaptations to group conversations (see Section~\ref{sec:discussion}).}

\wordingrevision{In adapting the framework, we also need to account for the asymmetric nature of the setting, since lawyers and justices occupy very different roles with different expectations for their conduct.}
Indeed, on average, lawyers take up more than twice as much talk-time as justices (median overall conversation imbalance of 0.74).
\wordingrevision{To reflect the institutionalized nature of these speaker roles, we by default assign lawyers and justices to the primary and secondary speaker roles respectively (\blue and \red in our visualization).
We additionally modify our criteria for determining which speaker dominates each conversation-window: lawyers need to talk more than $M_L=80\%$ of the speaking time to be considered as dominating that window (\blue in our visualization), while justices only need to talk more than $M_J=40\%$ to be considered as dominating the window (\red in our visualization).
If neither of these thresholds is reached, none of the parties are considered to dominate the window (\gray in the visualization). 
We also modify our criteria for the three stereotypical conversations from Figure 5:
we take back-and-forth conversations to be those with more than $40\%$ \gray windows,
alternating dominance conversations as those with than $40\%$ \red windows,
and dominating throughout conversations as those with more than $70\%$ \blue windows.}\footnote{\revision{In addition, we adjust the sliding window parameters $K = 120s$ and $L = 30s$ to account for 
shorter justices' talk-time in local spans. Minor variations of these parameters lead to similar results.}}

\begin{figure}[h]
  \centering
  \includegraphics[width=0.6\linewidth]{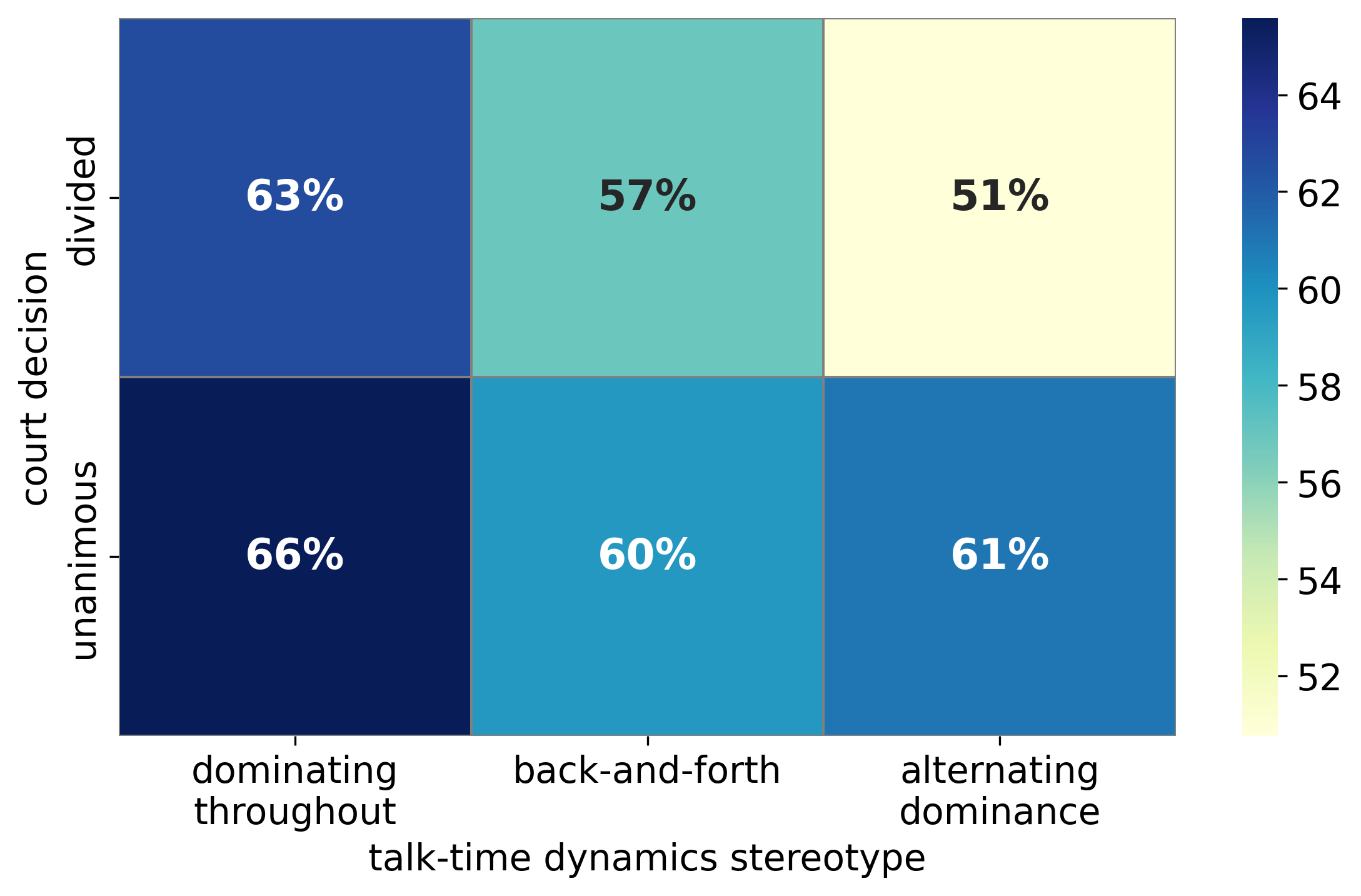}
  \caption{\wordingrevision{Comparing the share of cases won by petitioners across different conversation stereotypes and court decision types.}}
  \label{fig:supreme-court-convo1-heatmap}
\end{figure}

\xhdr{Case outcome and talk-time sharing dynamics}
Figure \ref{fig:supreme-court-convo1-heatmap} shows the
proportion of cases won by the petitioner, split up by whether the justices' votes were unanimous or divided, and by the talk-time sharing dynamics of the oral argument.
\wordingrevision{In cases where petitioners dominate the oral argument throughout, their share of wins is higher.}
Alternating dominance dynamics correspond to lower proportions of wins, perhaps reflecting more substantial scrutiny from the justices \cite{shapiro_oral_1984,jacobi_justice_2017}.
These differences are particularly notable in \wordingrevision{divided} cases, where perhaps more weight is placed on the petitioners' argumentation \cite{clark_politics_2022}.
\wordingrevision{These findings suggest that the conversational dynamics that our framework surfaces connect to the discursive and judicial processes taking place in Supreme Court cases. 
More broadly, they provide a starting illustration of how our framework can be applied in new settings, with appropriate modifications.}

%% file: 070conclusions.tex
In this work, we introduce a framework to model the dynamic distribution of talk-time in a conversation.
Starting from a basic account of conversation-level (im)balance between more or less talkative speakers,
our framework goes on to make finer-grained distinctions between various lower-level dynamics. 
We derive a structured space of talk-time sharing dynamics bounded by three stereotypical conversation patterns---conversations where one speaker continuously dominates the talk-time, conversations where two speakers alternate in dominating the talk-time, and back-and-forth conversations where neither speaker fully takes the lead.
We show, via a case study of a collection of video chats, how this perspective can enrich past accounts.
In particular, while we replicate existing findings that speakers generally enjoy balanced conversations (e.g., \citep{guydish_reciprocity_2021}),
we also find that even among fairly balanced conversations, 
speakers report very different experiences, 
depending on the particular dynamics of talk-time sharing involved.

\xhdr{Adapting to other modes of communication and communication settings}
\wordingrevision{While our main focus is on a specific conversational scenario---dyadic, synchronous, and role-neutral---future work could explore how talk-time dynamics play out in different scenarios.
This might require adapting some of the framework's basic elements, as demonstrated in our exploration of Supreme Court Oral Arguments in Section~\ref{sec:extension}.
For example, while in dyadic conversations imbalance can straightforwardly be measured as a fraction of talk-time, this would not be suitable for group conversations involving more than two parties.
While we used simple heuristics to treat the Supreme Court interactions as effectively dyadic,
future work could consider more sophisticated approaches, such as entropy-based measures previously used to measure balance in group conversations \cite{niculae_conversational_2016}.
In contexts where speakers have pre-determined roles that strongly condition talk-time sharing dynamics---such as in the Supreme Court Oral Arguments---those roles could replace or be added to the \primary and \secondary speaker distinction; 
this could help to more accurately model intuitive judgements of talk-time dominance, and also enable investigations of departures from expected dynamics.
Finally, it would be impossible to measure talk-time based on the time duration of utterances in written interactions;
simple adaptations could instead use the number of words (or even drafting-time) of an utterance instead.}
More generally, such simple extensions could expand the impact of the proposed framework, and serve as an additional basis for comparing interactions across different communication media \cite{scholl_comparison_2006,berger_communication_2013}.

The structured space of \TTsharing dynamics could be used to compare across different contexts.
For example,  one could also add to accounts of cultural variation in conversational practices \citep{tannen_indirectness_1981,scollon_athabaskan-english_1991,cook-gumperz_social_2006,dingemanse_text_2022}, going beyond the American setting considered here.
One could also examine the relation to social relationships between participants.
For instance, we find a dispreference for back-and-forths; this is somewhat surprising given past work linking such exchanges to feeling connected to one's conversational partner \citep{templeton_fast_2022}.
We speculate that in other settings, where conversational partners are closer to each other, such dynamics might point to flowing banter facilitated by shared understandings (e.g., \citep{wilkes-gibbs_coordinating_1992}); in our setting, which involves interactions between strangers, such a dynamic instead seems to point to an awkward search for conversation topics (as indicated by speakers' own accounts).

\xhdr{Examining linguistic processes}
Our approach could be enriched by accounting for the language used within conversations.
While our study used post-hoc reports from speakers to make sense of the various talk-time sharing dynamics we surfaced,
future work could examine what was said within and across different regimes of talk.
Here, we could draw on \citet{gilmartin_chats_2018}, who proposes an ontology to label different segments of talk (e.g., as storytelling, or as gossiping),
or on \citet{nguyen_modeling_2014}, who track the progression of topics in a conversation.

How do speakers' actions shape the talk-time sharing dynamics observed? 
Past work points to the various interactional moves that people can make in negotiating who has the floor, who should take the next turn, and when to switch \citep[inter alia]{sacks_simplest_1974,jefferson_caveat_1993,edelsky_whos_1981}.
Our method could be paired with such finer-grained approaches:
for instance, in tandem with locating conversations with alternating dominance patterns,
we could examine the linguistic or interactional phenomena present in particular cases
that might account for switches in lead speaker.

\xhdr{Design implications}
The \TTsharing framework has 
\wordingrevision{several potential applications relating to the}
design of computer mediated communication platforms, for both human-human and human-AI communication.
First, it can be used as a diagnostic tool:
platform maintainers could analyze interactions and their talk-time sharing dynamics,
drawing connections to inequities or to particular goals like participants' enjoyment or discussion constructiveness.
This could help inform ways of improving consequential processes like collaborative production \cite{lam_wpclubhouse_2011,gallus_gender_2020,halfaker_rise_2013}, participatory governance \cite{park_corpus_2018,frey_this_2019}, or the provision of care \cite{de_choudhury_gender_2017}.

Second, we suggest that designers who focused on computer-mediated communication 
settings should attend to the dynamics of talk-time sharing.
A number of past studies have proposed interfaces or automated tools to improve interactions, in part by making participants aware of the relative amount they're contributing \citep{viegas_chat_1999,leshed_visualizing_2009,do_how_2022,kim_meeting_2008}.
Our findings illustrate that it is important to distinguish between types of interaction that appear to have the same \overallimbalance of talk,
since different dynamics reflect diverging levels of speakers' enjoyment.
Other work points to imbalances in participation related to factors like gender and age \citep{lam_wpclubhouse_2011,gallus_gender_2020},
which can adversely impact collaborative projects.
We surface ways that such inequities can relate to finer-grained interaction patterns as well.

Third, our framework could help to more extensively explore design spaces;
for instance, interfaces could depict more complex interactional patterns such as the ones we've highlighted,
while interventions aimed at mediating discussions could better account for the dynamics of talk to improve their efficacy while minimizing their disruptiveness.
\revision{These ideas could enrich present approaches aimed at 
improving the treatment of newcomers to a platform \citep{halfaker_nice_2011}, 
alleviating social anxiety \citep{k_miller_meeting_2021},
encouraging empathy toward victims of cyberbullying \citep{taylor_accountability_2019},
diffusing tensions \citep{chang_thread_2022},
increasing constructiveness in teams \citep{niculae_conversational_2016,cao_my_2021},
\wordingrevision{alongside many other avenues} \citep{seering_designing_2019}.}
Future work could use our framework to design UI interventions that would inform speakers of ongoing talk-time sharing dynamics---e.g., by displaying a version of our visualization.
We note that such interventions depend on ways of more rigorously relating talk-time sharing dynamics to desired outcomes,
requiring careful user studies or causal analyses (e.g., see \citep{zhang_quantifying_2020}).

Finally, our work could inform the design and evaluation of conversational agents---an especially pertinent focus, given recent developments in AI framed around conversational interfaces, like ChatGPT.
A range of existing work has pointed to ways that people expect AI agents to behave from a conversation behavior standpoint \citep{zhang_ideal_2021}
and ways in which they fall short of realistic and fluent interaction \citep{reeves_conversational_2017,stokoe_elizabeth_2024,dingemanse_text_2022},
suggesting that careful attention to conversational mechanics can help with diagnosing existing issues
while calibrating future expectations.
Our framework can help with these efforts, in making the dynamic distribution of talk-time available for analysis.
Additionally, in revealing and making distinctions within a space of possible dynamics,
we add to accounts (e.g., \citep{dingemanse_text_2022})
that urge designers seeking to automate interaction to contend with the diversity of human conversation.

%% file: 120acknowledgements.tex
We are grateful for insightful discussions with Team Zissou---including Jonathan P. Chang, Yash Chatha, Nicholas Chernogor, Tushaar Gangavarapu, Kassandra Jordan, Seoyeon Julie Jeong, Sang Min Jung, Lillian Lee, Vivian Nguyen, Luke Tao, Son Tran, and Ethan Xia---and for the feedback we received from the anonymous reviewers.
We would also like to express our gratitude to the researchers from the BetterUp Labs for collecting and sharing the data, and to the CANDOR participants who generously shared their experiences for research purposes.
The first author would like to thank Yingchun Liu, Jibing Zhang, and Anastasia He for the support and helpful discussions.
Cristian Danescu-Niculescu-Mizil was funded in part by the U.S. National Science Foundation under Grant No. IIS-1750615 (CAREER).
Any opinions, findings, and conclusions in this work are those of the author(s) and do not necessarily reflect the views of Cornell University or the National Science Foundation.

%% file: 090appendixsection.tex
\subsection{Speaker Perception Across Axes}
\begin{figure}[h]
  \centering
  \includegraphics[width=0.6\linewidth]{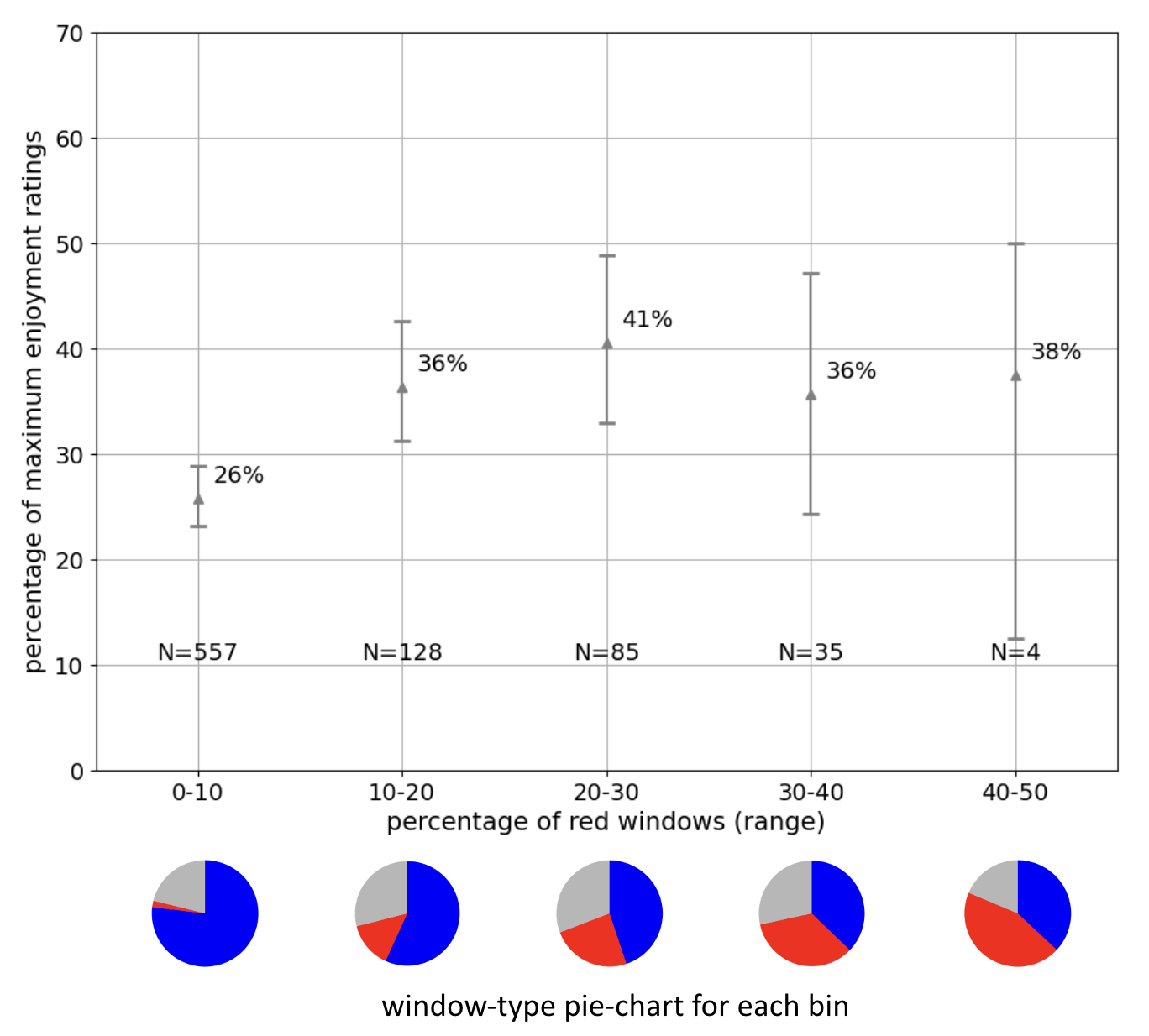}
  \caption{Similar to Figure~\ref{fig:candor_fix_blue_enjoy}, with similar percentage of windows that are neutral (\gray $<5\%$), we show difference in speaker perception between conversations with more windows dominated by the secondary speaker (\red) and conversations with more windows dominated by the primary speaker (\blue).
  }
  \Description{We show in plot more windows dominated by secondary speaker leads to higher percentage of speakers reporting maximum enjoyment rating, with number of gray windows controlled to be at the same level.}
  \label{fig:candor_fix_gray_enjoy}
\end{figure}

\begin{figure}[h]
  \centering
  \includegraphics[width=0.6\linewidth]{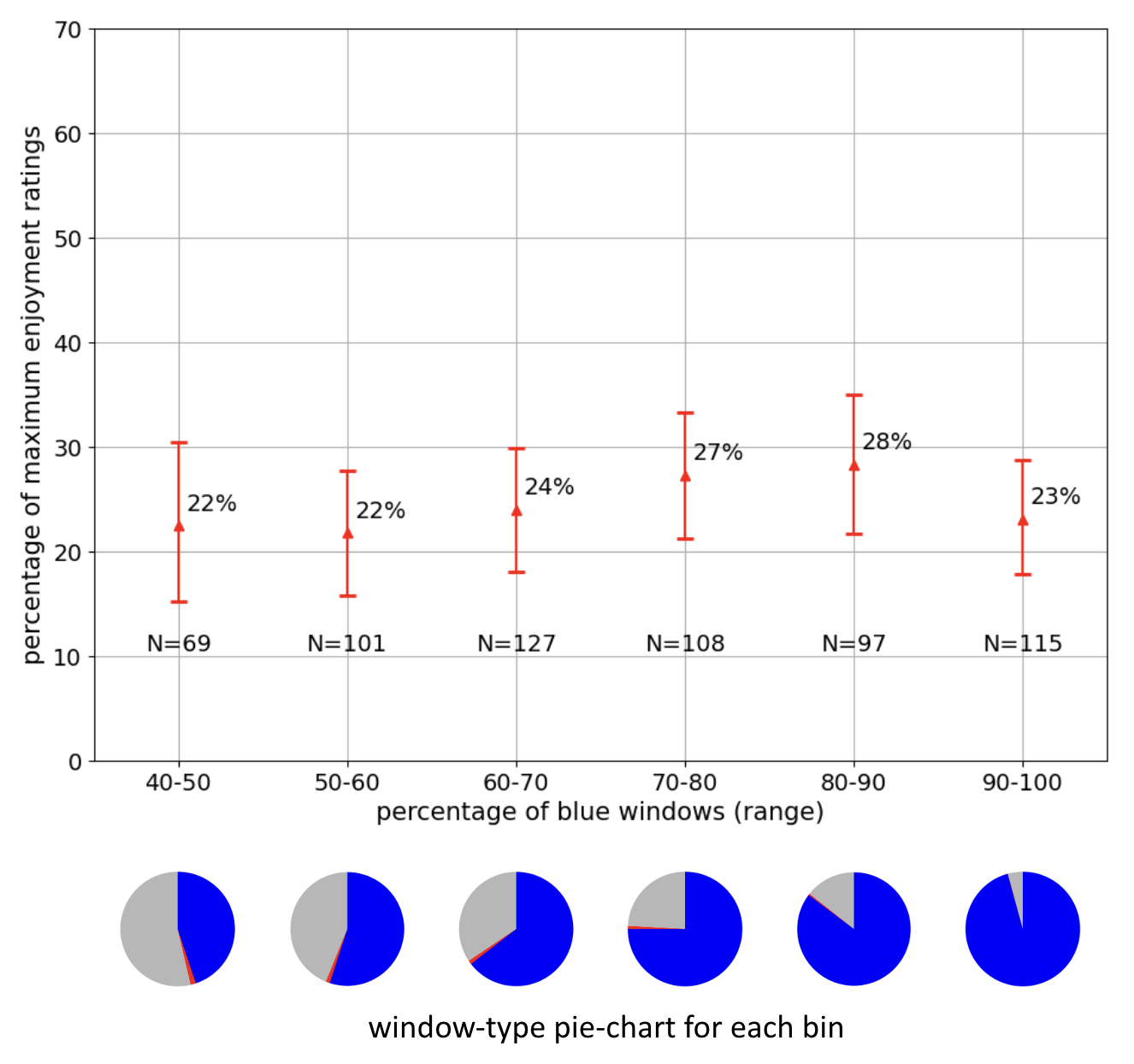}
  \caption{Similar to Figure~\ref{fig:candor_fix_blue_enjoy}, with similar percentage of windows dominated by the secondary speaker (\red $<5\%$), we show difference in speaker perception between conversations with more windows in neutral (\gray) and conversations with more windows dominated by the primary speaker (\blue).}
  \Description{We show in plot, with almost fixed percentage of windows dominated by secondary speaker (red) across all levels, the number of windows dominated by primary speaker (blue) has little correlation with enjoyment rating.}
  \label{fig:candor_fix_red_enjoy}
\end{figure}

We analyze the variations in enjoyment rating reported by speakers across the \blue-\red and \blue-\gray axis through controlling the amount of \gray windows to be below 40\% and the amount of \red windows to be below 5\% respectively. The results are presented in Figure~\ref{fig:candor_fix_gray_enjoy} and Figure~\ref{fig:candor_fix_red_enjoy}.
We observe limited difference in speaker perception along the \blue-\gray axis, and the result for the \blue-\red axis are similar to that for the \gray-\red axis shown in Figure~\ref{fig:candor_fix_blue_enjoy}.

\subsection{Age and Gender}
We present the distribution of conversations with varying \TTsharing dynamics across people from different demographic groups, specifically in terms of age and gender. For each \TTsharing dynamic stereotype, we show the percentage of conversations that occurs between two males, two females, or a male and a female (Table~\ref{tab:candor_gender_table}). A similar analysis is conducted for age groups, examining conversations between two younger speakers, two older speakers, and a younger and an older speaker (Table~\ref{tab:candor_age_table}). Younger and older speakers are classified based on the median age of all participants, which is 31 years old.
\begin{table*}
  \caption{Percentage of conversations with speaker from certain gender group (male, female, or others) across different \TTsharing dynamic stereotypes.}
  \label{tab:candor_gender_table}
  {\small
  \begin{tabular}{p{2cm}|p{2.5cm}|p{2cm}|p{2cm}|p{2cm}}
    \toprule
        & male and female & both male & both female & others \\ \hline
    overall\newline distribution &  48.9\% & 19.8\% & 28.0\% & 3.3\% \\ \hline
    dominating\newline throughout & 49.2\% & 20.7\% & 27.1\% & 3.0\% \\ \hline
    back and\newline forth & 53.4\% & 19.5\% & 24.3\% & 2.8\% \\ \hline
    alternating\newline dominance & 45.8\% & 18.8\% & 30.2\% & 5.2\% \\
  \bottomrule
\end{tabular}
} %
\end{table*}
\begin{table*}
  \caption{Percentage of conversations with speaker from certain age group (younger than 31 years old or older than 31 years old) across different \TTsharing dynamic stereotypes.}
  \label{tab:candor_age_table}
  {\small
  \begin{tabular}{p{2cm}|p{3cm}|p{2.5cm}|p{2.5cm}}
    \toprule
        & younger and older & both younger & both older \\ \hline
    overall\newline distribution &  51.1\% & 21.1\% & 27.9\% \\ \hline
    dominating\newline throughout & 56.3\% & 18.6\% & 25.1\% \\ \hline
    back and\newline forth & 50.2\% & 29.1\% & 20.7\% \\ \hline
    alternating\newline dominance & 53.1\% & 17.7\% & 29.2\% \\
  \bottomrule
\end{tabular}
} %
\end{table*}

\subsection{Measuring speak-time from audio vs. transcripts}
\label{sec:appendix-audio}
\begin{figure}[h]
  \centering
  \includegraphics[width=0.9\linewidth]{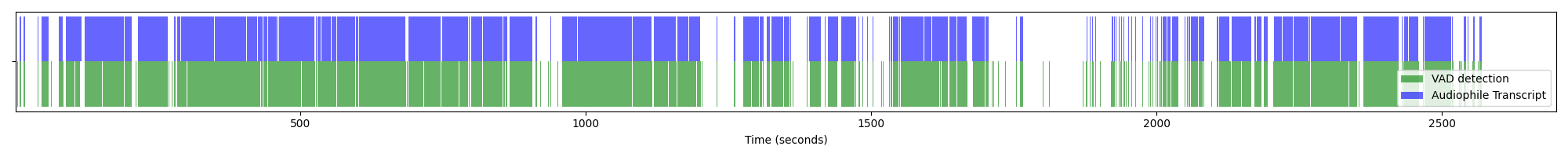}
  \caption{Visualization on the alignment of timestamps between automated transcript timestamps and voice activity detection results for an individual conversation speaker audio file (CANDOR ID: 68e5b791-3931-440e-a603-4e3a063d8521, speaker ID: 5f880d491c31fc25c442fa70). For most of the CANDOR audio files, we observe good alignment to validate the accuracy of automated transcript timestamps.}
  \Description{An visualization on the alignment of timestamps. The plot show good alignment between the two, with each speaking activity start and end points aligning well.}
  \label{fig:candor_good_audio_vad}
\end{figure}

\begin{figure}[h]
  \centering
  \includegraphics[width=0.9\linewidth]{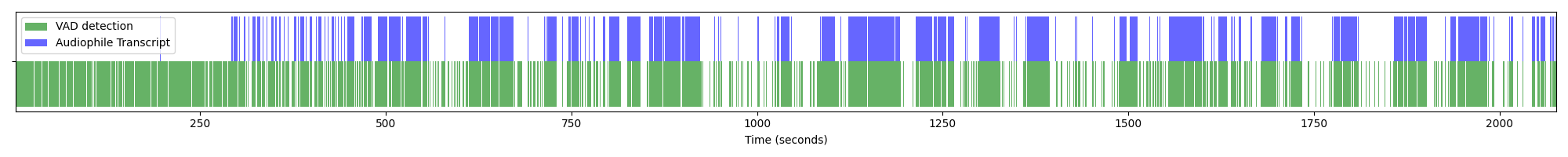}
  \caption{Visualization of a conversation audio file with bad alignment of timestamps between automated transcript timestamps and voice activity detection results (CANDOR ID: dbbf2d8f-eaba-478c-bf32-330620a4b4ee, speaker ID: 5a70d87e9cdd180001776440). The error in poorly aligned timestamps is hypothesized to be with the problem of voice activity detection's inability to distinguish between speaking language and noise or backchannel voices, which results in extra detected speaking time at the beginning of the conversation that are likely to be noise in speaker's waiting time for this audio.}
  \Description{An visualization of a poorly aligned timestamp between CANDOR automated transcript and voice activity detection result. voice activity detection produces long sequences of extra speaking times at the beginning of the conversation that are likely to be noise in speaker's waiting time.}
  \label{fig:candor_bad_audio_vad}
\end{figure}

\revision{
We present the validation of transcript accuracy in capturing speak-time by comparing it against extracted timestamps that we extract from the raw audio recordings using a voice activity detection tool, specifically the webrtcvad package in python.
We consider the speaking time overlaps between automated transcript and voice activity detection result for every individual conversation speaker.
Among all 1594 conversations we used in our analysis (3188 audio files, 2 for each conversation separated by speakers), we observe a median f1 score of 0.86 (0.93 precision and 0.81 recall) across all speaker audio files, with only 6.88\% having f1 score less than 0.7.
Figure~\ref{fig:candor_good_audio_vad} and figure~\ref{fig:candor_bad_audio_vad} present two visualizations of the alignment between CANDOR's automated transcript timestamps and the voice activity detection results for the audio recordings of two conversation speakers.
We observe a high alignment between the two for most conversations, such as the one shown in Figure~\ref{fig:candor_good_audio_vad}, particularly in terms of the true positive rate (resulting in a high precision score as mentioned above).
However, in rare cases we also notice that additional speech activities detected from the audio signal, as shown in Figure~\ref{fig:candor_bad_audio_vad}.
Based on a manual investigation of these cases, these differences can be largely attributed to noise or non-speech sounds, such as laughter or coughing.
This results in a lower recall comparing to precision score across audio files.
Nevertheless, these "noise" signals could provide additional information to support future analyses. For example, the frequency of laughter could be used to infer participants' experiences during the conversation.
Future work could explore additional uses of the audio files to leverage richer signals within.}

\begin{figure}[h]
  \centering
  \includegraphics[width=0.6\linewidth]{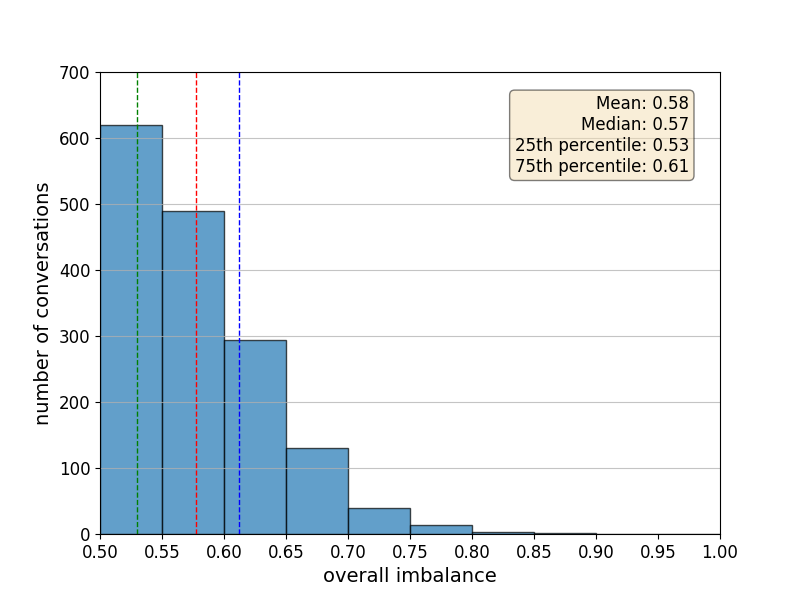}
  \caption{Distribution of conversation-level imbalance in the CANDOR corpus with timestamps directly extracted from raw audio files using voice activity detection tool.}
  \Description{the distribution of overall imbalance shift more to the right comparing to figure~\ref{fig:candor_balance_distribution}, with mean imbalance as 0.58, median 0.57, 25th and 75th percentile 0.53 and 0.61.}
  \label{fig:candor_audio_overall_balance}
\end{figure}

\begin{figure}[h]
  \centering
  \includegraphics[width=0.6\linewidth]{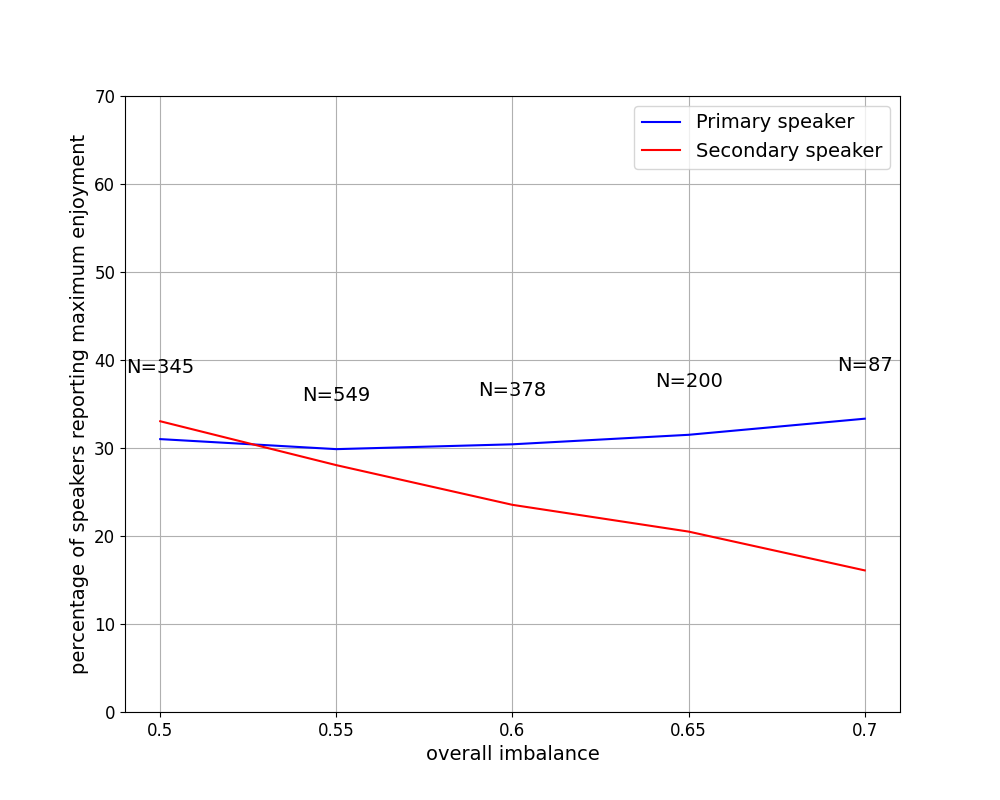}
  \caption{Similar to Figure~\ref{fig:candor_enjoy_vs_balance} but with speaking timestamps captured using voice activity detection tool, we show the percentage of primary and secondary speakers reporting meximum enjoyment scores for conversations with different levels of imbalance. As in Figure~\ref{fig:candor_enjoy_vs_balance}, levels with 30 or less conversations ($N < 30$) are not shown. The same qualitative trend is observed.}
  \Description{Same qualitative trend obeserved as Figure~\ref{fig:candor_enjoy_vs_balance}, with more conversations in the more balanced side.}
  \label{fig:candor_audio_speaker_enjoy}
\end{figure}

\begin{figure}[h]
  \centering
  \includegraphics[width=0.58\linewidth]{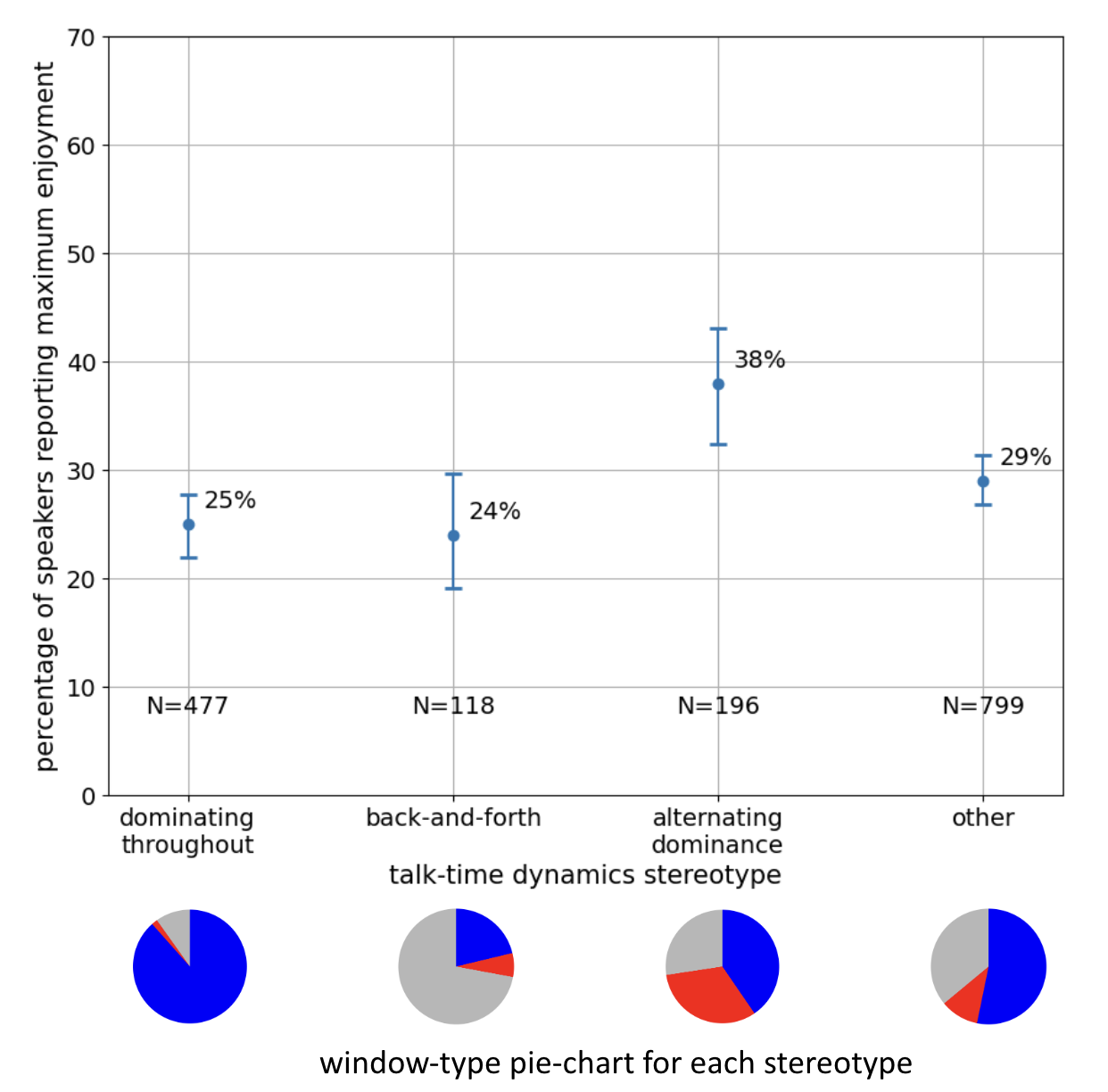}
  \caption{Similar to Figure~\ref{fig:candor_enjoy_triangle} but with speaking timestamps captured using voice activity detection tool, we compare the reported enjoyment of the three stereotypical talk-time dynamics: dominating throughout, back-and-forth, and alternating dominance. The same qualitative trend is observed.}
  \Description{Percentage of speakers reporting maximum enjoyment is highest for alternating dominance type conversations, followed by dominating throughout and back-and-forth is lowest. The error bar was larger for alternating dominance and back-and-forth due to less datapoints, but the order remains the same. Pie charts are also presented fro each type, with mostly blue for dominating throughout, mostly gray for back-and-forth, and almost equal for all three colors for alternating dominance.}
  \label{fig:candor_audio_talk_time_dynamics_enjoy}
\end{figure}

\begin{figure}[h]
  \centering
  \includegraphics[width=0.58\linewidth]{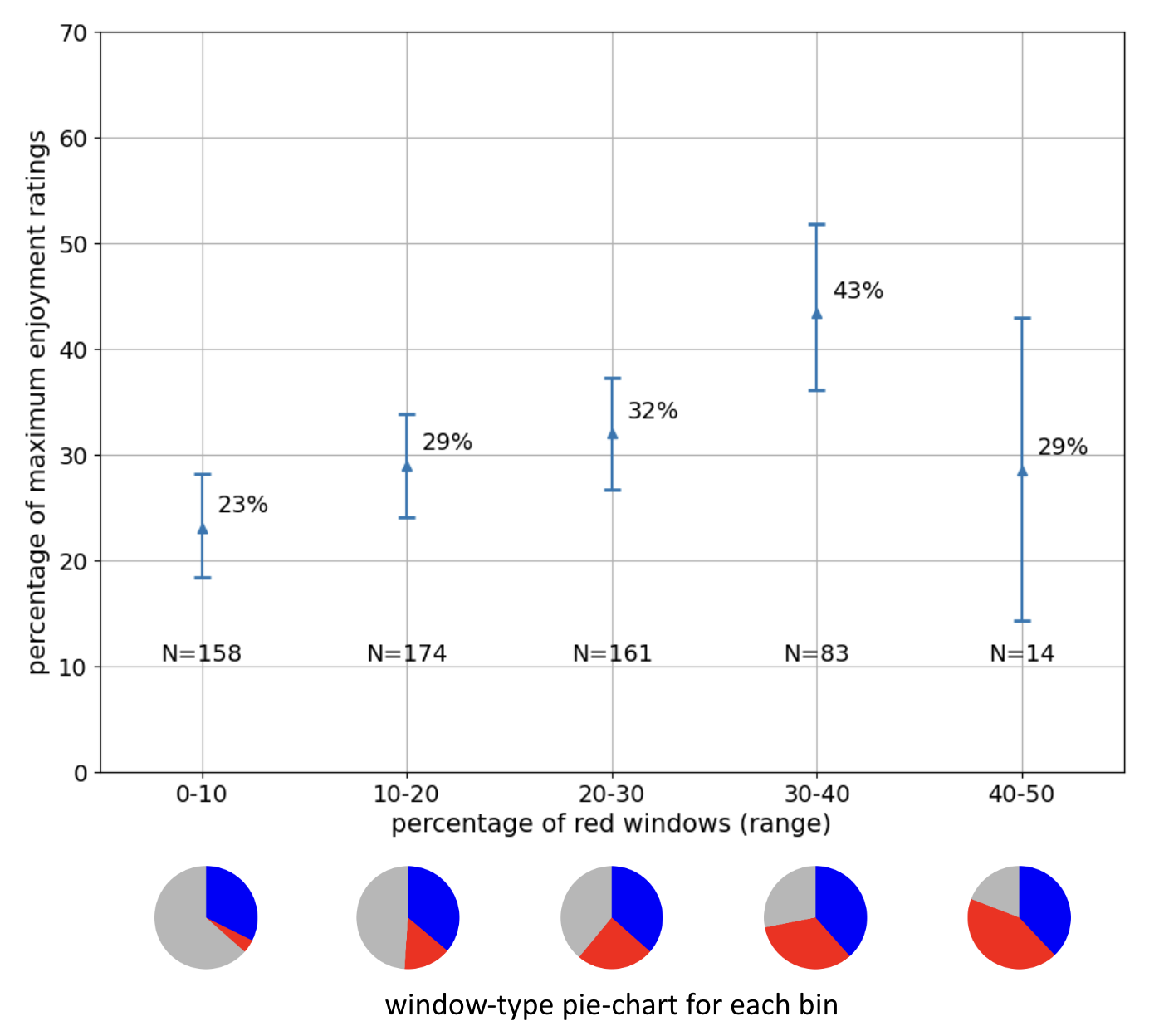}
  \caption{Similar to Figure~\ref{fig:candor_fix_blue_enjoy} but with speaking timestamps captured using voice activity detection tool, we control for the proportion of \blue windows, and compare reported enjoyment for varying proportions of \red windows. The same qualitative trend is observed.}
  \Description{We show in plot, with almost fixed percentage of windows dominated by primary speaker (blue) across all levels, the more windows dominated by secondary speaker (red), the higher enjoyment perceived by participants.}
  \label{fig:candor_audio_along_axis}
\end{figure}

\revision{
To validate our results from the exploration of CANDOR corpus, we conduct all analysis again directly on the audio files with voice activity detection algorithm extracted timestamps.
We observe same qualitative trend from our main analysis, with slightly more balanced talk-time sharing behavior across conversation as shown in Figure~\ref{fig:candor_audio_overall_balance}.
This slight change could be the result of the extra non-language activities discussed above.
We take this into account by adjusting the window-dominance parameter accordingly ($M = 55\%$) and present the results in Figure~\ref{fig:candor_audio_speaker_enjoy}, ~\ref{fig:candor_audio_talk_time_dynamics_enjoy}, and ~\ref{fig:candor_audio_along_axis}.}

\subsection{Parameters for Supreme Court Oral Arguments and additional results}
\label{sec:appendix-supreme-court}

\begin{figure}[h]
  \centering
  \includegraphics[width=0.6\linewidth]{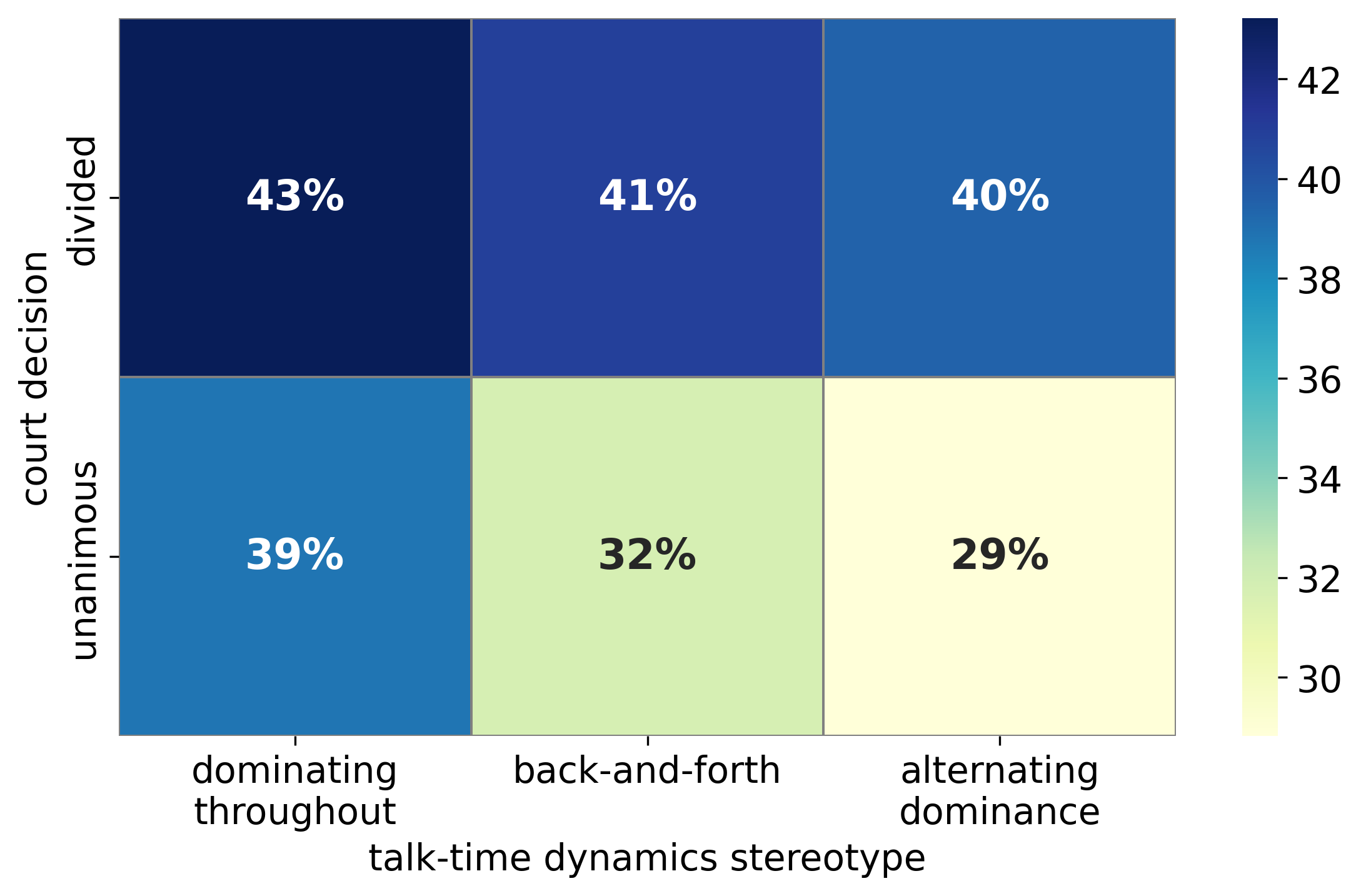}
    \caption{Comparing the respondent' winning rate across different stereotypes of conversations and in different court decision types.}
  \label{fig:supreme-court-convo2-heatmap}
\end{figure}

\xhdr{Respondent side results}
\label{sec:supreme-respondent}
\revision{
Here we present results from respondent's turn of the oral arguments.
As shown in Figure \ref{fig:supreme-court-convo2-heatmap}, a similar trend is observed. When lawyers dominate throughout oral arguments, they have a greater chance of winning; otherwise, they are more likely to lose the case.  
The difference is more pronounced in unanimous cases, in which the respondents are substantially less likely to win.
}

\subsection{Additional Examples}

\revision{
In this section, we present additional visualizations and examples of speaker comments to further support our qualitative interpretation.}

\begin{figure}[h]
  \centering
  \includegraphics[width=0.7\linewidth]{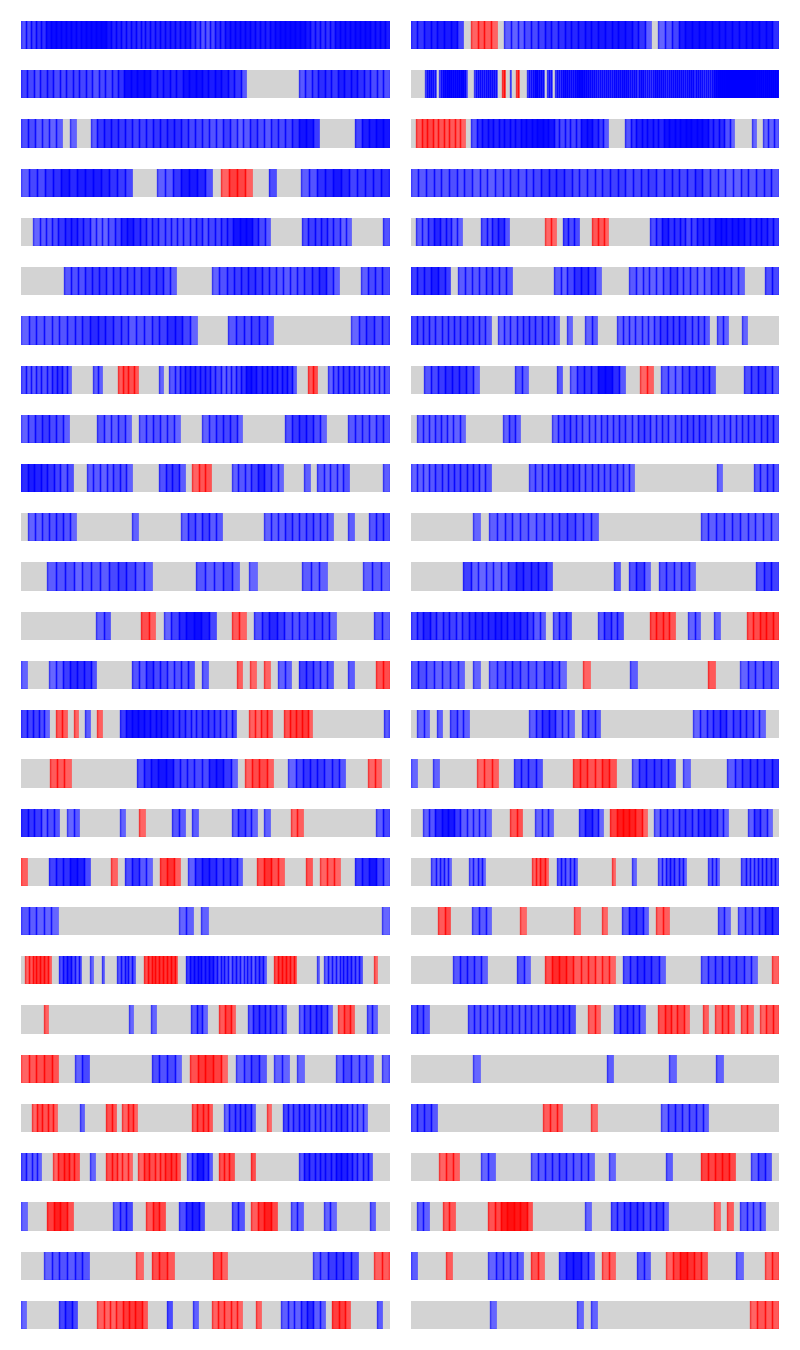}
  \caption{Visualization of talk-time dynamics of 54 conversations in the CANDOR dataset, sorted in descending conversation level imbalance.}
  \Description{54 conversation talk-time dynamics in CANDOR, with top half of the conversations largely covered in blue, and bottom half of the conversations having mixed of more gray or more red.}
  \label{fig:multi-convo-sort-balance}
\end{figure}

\xhdr{More conversation visualization showcases}
\revision{
We present more conversation talk-time dynamics visualization from conversations in the CANDOR corpus. Figure~\ref{fig:multi-convo-sort-balance} shows the visualization of talk-time dynamics for 54 conversations sorted by their conversation-level imbalance.
Note that while more imbalance conversations are largely in dominating throughout stereotype, more balanced conversations can be either back-and-forth or alternating dominance.}

\begin{figure}[h]
  \centering
  \includegraphics[width=0.7\linewidth]{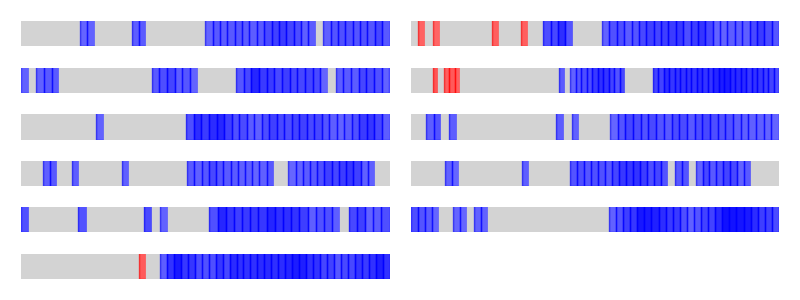}
  \caption{Back-and-forth transitioning to dominating throughout conversations}
  \Description{for these conversations, first half of the conversation are mostly gray, second half mostly blue.}
  \label{fig:bf-dt}
\end{figure}

\begin{figure}[h]
  \centering
  \includegraphics[width=0.7\linewidth]{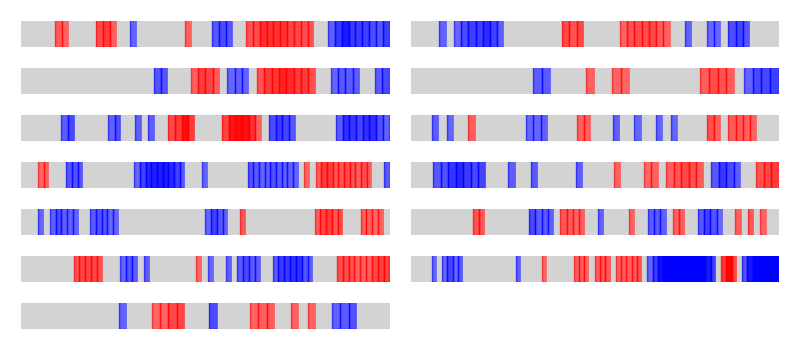}
  \caption{Back-and-forth transitioning to alternating dominance conversations}
  \Description{for these conversations, first half of the conversation are mostly gray, second half have more red in it.}
  \label{fig:bf-ad}
\end{figure}

\begin{figure}[h]
  \centering
  \includegraphics[width=0.7\linewidth]{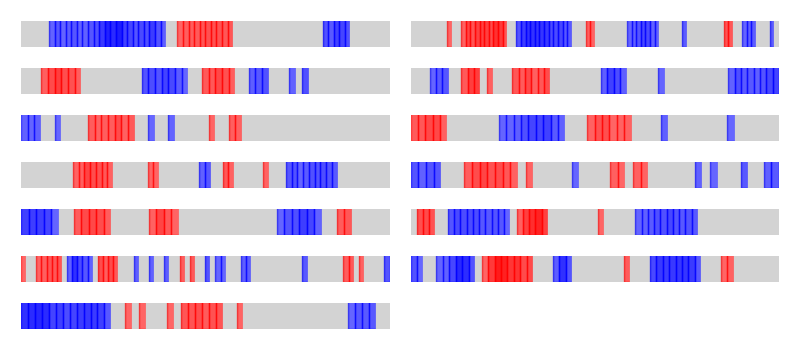}
  \caption{Alternating dominance transitioning to back-and-forth conversations}
  \Description{for these conversations, first half of the conversation have more red in it, second half are mostly gray.}
  \label{fig:ad-bf}
\end{figure}

\begin{figure}[h]
  \centering
  \includegraphics[width=0.7\linewidth]{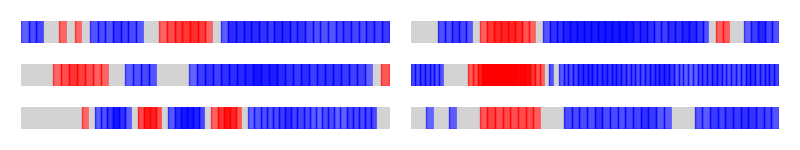}
  \caption{Alternating dominance transitioning to dominating throughout conversations}
  \Description{for these conversations, first half of the conversation have more red in it, second half are mostly blue.}
  \label{fig:ad-dt}
\end{figure}

\begin{figure}[h]
  \centering
  \includegraphics[width=0.7\linewidth]{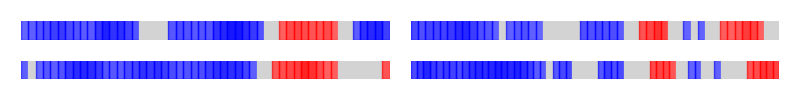}
  \caption{Dominating throughout transitioning to alternating dominance conversations}
  \Description{for these conversations, first half of the conversation are mostly blue, second half have more red in it.}
  \label{fig:dt-ad}
\end{figure}

\xhdr{Combined stereotype conversations visualization showcases}
\revision{
We present visualization from conversations with a combination of stereotypes as described in Section~\ref{sec:method}.
Figures~\ref{fig:bf-dt}, ~\ref{fig:bf-ad}, ~\ref{fig:ad-bf}, ~\ref{fig:ad-dt}, and ~\ref{fig:dt-ad} shows all 46 conversations with a combination of stereotypes captured by our framework.}

\subsubsection{More User Comment Examples}

\revision{
In this section, we provide expanded versions of Table~\ref{tab:candor_fw_balance_imbalance}, ~\ref{tab:candor_fw_primary_secondary}, and ~\ref{tab:candor_fw_alternate_backforth} including the full speaker comments to give a more complete picture of how users describe their experiences. See Table~\ref{tab:candor_fw_balance_imbalance_extended_a}, ~\ref{tab:candor_fw_balance_imbalance_extended_b}, ~\ref{tab:candor_fw_primary_secondary_extended_a},
~\ref{tab:candor_fw_primary_secondary_extended_b}, 
~\ref{tab:candor_fw_alternate_backforth_extended_a}, and ~\ref{tab:candor_fw_alternate_backforth_extended_b}.}

\begin{table*}[t]
  \caption{Extended version of Table~\ref{tab:candor_fw_balance_imbalance}: phrases from positive comments that most distinguish between \balanced and \imbalanced conversations, with representative excerpts.}
  \label{tab:candor_fw_balance_imbalance_extended_a}
  {\small
  \begin{tabular}{p{3.3cm}|p{2.2cm}|p{7cm}}
    \toprule
            & top distinctive phrases & example excerpts \\ \hline
    Positive~comments~for \textbf{\highlybalanced} \newline conversations & 
    both, lot in common, were able, were able to, like we, lot in, about our, able, able to, we both &
    \noindent \textbullet \textbf{We both} had a \textbf{lot in common}, so it made talking with each other super easy, and almost \textbf{like we} knew each other. \newline
    \textbullet\ We had a lot to talk about, and seemed to have a \textbf{lot in common}.
\textbf{We both} had single children and \textbf{were able to} connect with that. \newline
    \textbullet\ We \textbf{were able to} connect well over our experiences with prolific and the study as well as sharing similar hobbies and likes. \newline
    \textbullet\ I felt \textbf{like we} listened to one another. \newline
    \textbullet\ I feel \textbf{like we} connected and that makes it easier. \newline
    \textbullet\ It was interesting learning \textbf{about our} similarities and differences in ideas. \newline
    \textbullet\ I felt open and \textbf{able to} share what is going on in my city and life. \\ \hline
    
    Positive comments for \textbf{\highlyimbalanced} conversations & 
    chat, he, partner, talked lot, listener, else, im, is, she, bring &
    \noindent \textbullet\ I feel like my \textbf{chat partner} was great at asking me questions and keeping me talking. \newline
    \textbullet\ It seems like \textbf{he} is a talker in real life and \textbf{he} likes to be the center of attention, so maybe this helped in our video \textbf{chat} that we had. \newline
    \textbullet\ \textbf{She talked a lot} and was very pleasent and smiled a lot. \newline
    \textbullet\ Although I tend to be an introvert, I know a \textbf{talked a lot} and had to intentionally stop myself and ask him questions and invite him to talk more. \newline
    \textbullet\ \textbf{She} was also relaxed, an active \textbf{listener}, and asked good questions. \newline
    \textbullet\ Fortunately, I am a good \textbf{listener}. \newline
    \textbullet\ I prefer when the other person \textbf{brings} up conversation topics because it is sometimes difficult for me to come up with things to say. \\
  \bottomrule
  \end{tabular}
  } %
\end{table*}

\begin{table*}[t]
  \caption{Extended version of Table~\ref{tab:candor_fw_balance_imbalance}: phrases from negative comments that most distinguish between \balanced and \imbalanced conversations, with representative excerpts.}
  \label{tab:candor_fw_balance_imbalance_extended_b}
  {\small
  \begin{tabular}{p{3.3cm}|p{2.2cm}|p{7cm}}
    \toprule
            & top distinctive phrases & example excerpts \\ \hline
    Negative~comments~for \textbf{\highlybalanced} \newline  conversations & 
    our, our conversation, think of, of anything that, we had, anything that, both, her and, think the conversation, each other &
    \noindent \textbullet\ Honestly, \textbf{our conversation} was top notched! We hit it off right away and had no issues reaching the 25 minute mark. \newline
    \textbullet\ We flowed easily in \textbf{our conversation} and I am typing to get to fifty word count. \newline
    \textbullet\ I think the rough parts were mainly how we tried to \textbf{think of} how to go back and forth, at first. \newline
    \textbullet\ I can’t really \textbf{think of} anything that went wrong. \newline
    \textbullet\ We \textbf{both} tend to be more introverted as we mentioned in the conversation. \newline
    \textbullet\ I felt like I was under pressure to just agree with \textbf{her and} nod along even when I was not interested. \newline
    \textbullet\ We spent twice the recommended time chatting and I could totally see being friends with her if we were closer to \textbf{each other}. \\ \hline
    
    Negative~comments for \textbf{\highlyimbalanced} conversations & 
    than, more than, much, talked, me, too much, talked too, talked too much, he, ended up, ended &
    \noindent \textbullet\ I think \textbf{he} did talk a fair amount \textbf{more than} me, which was mildly frustrating, but \textbf{he} meant well and I understood it better when \textbf{he} told me \textbf{he} was originally from New York. \newline
    \textbullet\ \textbf{He talked} a lot \textbf{more than} I did. \newline
    \textbullet\ I guess one thing was that she did most of the talking, but I was happy to listen to someone who knows \textbf{much more than} I do. \newline
    \textbullet\ I think I \textbf{talked too much}. \newline
    \textbullet\ I probably \textbf{talked too much}, but I felt like she wasn't giving very much in return so I don't think I had a choice. \newline
    \textbullet\ I think she felt that way too and we \textbf{ended up} sort of slowing down and stretching out to cover the time needed. \newline
    \textbullet\ \textbf{He ended up} opening up toward the end and actually provided me with some information that I might find useful for everyday life as well. \\
  \bottomrule
  \end{tabular}
  } %
\end{table*}

\begin{table*}[t]
  \caption{Extended version of Table~\ref{tab:candor_fw_primary_secondary}: phrases that most distinguish between positive comments by \primary and \secondary speakers for highly-\imbalanced conversations.}
  \label{tab:candor_fw_primary_secondary_extended_a}
  {\small
  \begin{tabular}{p{3cm}|p{2.4cm}|p{7cm}}
    \toprule
            & top distinctive phrases & example excerpts \\ \hline
    Positive comments \newline from \textbf{\primary speaker} & 
    questions, asked, but, probably, us, about my, agreed, some, young, good listener, each, interested &
    \noindent \textbullet She was a \textbf{good listener} and \textbf{asked} pertinent \textbf{questions} to get me talking. \newline
    \textbullet\ My high energy and inquisitive nature allowed me to ask many \textbf{questions} of the other participant. \newline
    \textbullet\ The way he was curious and listened and \textbf{asked} more and more \textbf{questions}. \newline
    \textbullet\ I think that both of \textbf{us} felt awkward at first, but she did a great job of normalizing that and she seemed to have a lot to talk about. \newline
    \textbullet\ I probably did the majority of the talking, but they \textbf{asked} good \textbf{questions} and gave interesting answers when I \textbf{asked} them \textbf{questions}. \newline
    \textbullet\ I was able to learn a lot \textbf{about my} partner without being prying on invasive. \newline
    \textbullet\ She was also fun to talk to because she's a \textbf{good listener} and was understanding toward my situation. \\ \hline

    Positive comments \newline from~\textbf{\secondary speaker} & 
    had lot, something, lot, stories, had similar, had lot of, share, friendly and, that my, life &
    \noindent \textbullet He \textbf{had a lot} to talk about so I let him tell me about those things for as long as he wanted to. \newline
    \textbullet\ We \textbf{had a lot} in common even though we were very different people. \newline
    \textbullet\ We found \textbf{something} that we could connect on right away. \newline
    \textbullet\ He had great life experience and \textbf{stories} that he \textbf{shared}. \newline
    \textbullet\ We also \textbf{had similar} experiences in some ways, which helped out. \newline
    \textbullet\ My partner was very chatty and willing to \textbf{share} her views, while I was quite willing to listen. \newline
    \textbullet\ We both were \textbf{friendly and} eager to listen to each other. \\
  \bottomrule
  \end{tabular}
  } %
\end{table*}

\begin{table*}[t]
  \caption{Extended version of Table~\ref{tab:candor_fw_primary_secondary}: phrases that most distinguish between negative comments by \primary and \secondary speakers for highly-\imbalanced conversations.}
  \label{tab:candor_fw_primary_secondary_extended_b}
  {\small
  \begin{tabular}{p{3cm}|p{2.4cm}|p{7cm}}
    \toprule
            & top distinctive phrases & example excerpts \\ \hline
    Negative comments \newline from \textbf{\primary speaker} & 
    her, little, too much, we had, flow, talked too, should, she was, she is, shy &
    \noindent \textbullet My partner was not the best at \textbf{her} side of the discussion, sometimes failing to move the conversation forward and creating awkward silences. \newline
    \textbullet\ I felt a \textbf{little} scattered about my answers. \newline
    \textbullet\ I think \textbf{she was} a \textbf{little} quiet by nature which is ok. \newline
    \textbullet\ The conversation felt somewhat forced, \textbf{we had} to find things to talk about. \newline
    \textbullet\ I think I probably \textbf{talked too much}. \newline
    \textbullet\ I felt I led the conversation \textbf{too much} and perhaps dominated. \newline
    \textbullet\ He was kind of \textbf{shy}, and quiet. \newline
    \textbullet\ She seemed very distracted by something else, which made it difficult for the conversation to \textbf{flow}. \\ \hline

    Negative comments \newline from \textbf{\secondary speaker} & 
    different, hard, say, went, the conversation was, talked lot, life, honestly, didnt, kept, it was hard &
    \noindent \textbullet I think we are just two very \textbf{different} people, so we \textbf{didn't} have a lot in common. \newline
    \textbullet\ He talked way too much; \textbf{it was hard} to get a word in edgewise. \newline
    \textbullet\ I did not know what to \textbf{say}. \newline
    \textbullet\ What \textbf{didn't} work in \textbf{the conversation was} the lack of equity in conversation talking time. \newline
    \textbullet\ My partner \textbf{talked a lot} about herself and \textbf{didn't} ask many questions, which made it harder for me to share as well. \newline
    \textbullet\ He \textbf{talked a lot} more than I did. \newline
    \textbullet\ In real \textbf{life} I tend to be on the quiet side so this was expected. \\
  \bottomrule
  \end{tabular}
  } %
\end{table*}

\begin{table*}[t]
  \caption{Extended version of Table~\ref{tab:candor_fw_alternate_backforth}: phrases that most distinguish between positive comments for stereotypical alternating dominance and back-and-forth conversations.}
  \label{tab:candor_fw_alternate_backforth_extended_a}
  {\small
  \begin{tabular}{p{2.8cm}|p{2.6cm}|p{7cm}}
    \toprule
            & top distinctive phrases & example excerpts \\ \hline
    Positive~comments~for \textbf{alternating dominance} conversations & 
    stories, to listen, listen, stranger, life experiences, the other, myself, his, good conversation, super &
    \noindent \textbullet\ We both had interesting \textbf{stories} to tell, so that was also very enjoyable. \newline
    \textbullet\ She had some really interesting \textbf{life experiences} and \textbf{stories} to share. \newline
    \textbullet\ I feel like we were both able \textbf{to listen} actively to one another, and bring curiosity to the conversation. \newline
    \textbullet\ We both listened when \textbf{the other} was speaking and asked follow-up questions. \newline
    \textbullet\ I felt that that the conversation did not feel like I was talking to a complete \textbf{stranger}. \newline
    \textbullet\ I felt comfortable talking to \textbf{the other} person and was willing to open up about \textbf{myself}, which I think led to a \textbf{good conversation} overall. \newline
    \textbullet\ We had a \textbf{good conversation} about several topics about movies, books, and work. \\ \hline
    
    Positive comments for \textbf{back-and-forth} conversations & 
    which, the conversation going, conversation going, topics, keep the, keep the conversation, games, going, well was, discussed &
    \noindent \textbullet\ We both took turns talking and \textbf{keeping the conversation going} with each other. \newline
    \textbullet\ He \textbf{kept the conversation going} at a good pace, and really was able to steer the conversation back when I ventured off \textbf{topic}, or was obviously uncomfortable. \newline
    \textbullet\ The flow of the conversation went well in which we kept bringing up \textbf{topics} to speak on and it went smoothly. \newline
    \textbullet\ I think the wide variety of \textbf{topics} we \textbf{discussed} was interesting and informative.\newline
    \textbullet\ Not a lot outside of talking about \textbf{games}. \newline
    \textbullet\ So we \textbf{discussed} many of our favorite \textbf{games} together and talked about our emotional responses to the \textbf{games}. \newline
    \textbullet\ What worked \textbf{well was} that my partner was willing to keep talking. \\
  \bottomrule
  \end{tabular}
  } %
\end{table*}

\begin{table*}[t]
  \caption{Extended version of Table~\ref{tab:candor_fw_alternate_backforth}: phrases that most distinguish between negative comments for stereotypical alternating dominance and back-and-forth conversations.}
  \label{tab:candor_fw_alternate_backforth_extended_b}
  {\small
  \begin{tabular}{p{2.8cm}|p{2.6cm}|p{7cm}}
    \toprule
            & top distinctive phrases & example excerpts \\ \hline
    Negative comments for \textbf{alternating dominance} conversations & 
    go, one, interested, she, into the, minutes, get, wasnt really, through, few minutes &
    \noindent \textbullet\ I don't \textbf{really} have much to say about what didn't \textbf{go} well. \newline
    \textbullet\ My partner didn't seem \textbf{interested} in me at all. \newline
    \textbullet\ I hope that \textbf{she} felt able, and had the space to direct the conversation along lines that \textbf{interested} her. \newline
    \textbullet\ He did a good job listening to me when I talked, but sometimes he didn't easily let me \textbf{into the} conversation. \newline
    \textbullet\ As soon as we were a \textbf{few minutes} over, \textbf{she} wanted to end the conversation. \newline
    \textbullet\ I fell guarded the first \textbf{few minutes} of the conversation because I did not know how to proceed with this person. \newline
    \textbullet\ \textbf{She} laughed at stuff that \textbf{wasn't really} funny several times, so I couldn't tell if it was for humor or nerves. \\ \hline

    Negative comments for \textbf{back-and-forth} conversations & 
    some, going, im, most, had to, went well, understand, flow, kept, topics, moments &
    \noindent \textbullet\ The conversation had \textbf{some} awkward \textbf{moments} and silent pauses. \newline
    \textbullet\ Since we were both a bit more shy, we had \textbf{some moments} of dead air. \newline
    \textbullet\ There were \textbf{some moments} where I wasn't sure what to talk about next. \newline
    \textbullet\ \textbf{Most} of the conversation did work well, but I found I \textbf{had to} do most of the prompting at the start of the conversation. \newline
    \textbullet\ He didn't have a lot of talking points and I felt as though I \textbf{had to} keep asking questions to keep the conversation \textbf{going}. \newline
    \textbullet\ The conversation \textbf{flow} seemed forced especially when one topic had been discussed and we need to find another. \newline
    \textbullet\ The communication, I couldn't always \textbf{understand} him, as I said his accent was pretty strong for me to follow and always \textbf{understand}. \newline
    \textbullet\ There were times where it felt like I was interrupting what he was trying to say or I didn't fully grasp or \textbf{understand} what he was saying. \\
  \bottomrule
  \end{tabular}
  } %
\end{table*}